\documentclass[final]{siamltex}

\usepackage[]{fontenc}
\usepackage[latin1]{inputenc}
\usepackage{verbatim}
\usepackage{amsmath}
\usepackage{amssymb}
\usepackage{dsfont}
\usepackage{amsfonts}
\usepackage{mathrsfs}
\usepackage{graphicx}
\usepackage{color}
\usepackage{ulem}
\usepackage{setspace}
\usepackage{xspace}
\usepackage{ stmaryrd }
\usepackage{graphicx}
\usepackage{lineno}
\usepackage{ stmaryrd }

\DeclareMathOperator*{\argmin}{arg\,min}
\newcommand\xx{\mathbf{x}}

\newcommand{\zz}{\boldsymbol{z}}
\newcommand{\Zc}{\Xi}
\newcommand{\Bm}{\mathbf{B}}
\newcommand{\Am}{\mathbf{A}}
\newcommand{\Rb}{\mathbb{R}}
\newcommand{\Rr}{\mathbb{R}}
\newcommand{\uv}{\mathbf{u}}
\newcommand{\eg}{\textit{e.g.}}
\newcommand{\ie}{\textit{i.e.}}
\newcommand{\oo}{o}

\newcommand\yy{\mathbf{y}}

\def\wwr{\mbox{$\mathbf{{v}}^k$}}
\def\ww{\mbox{$\mathbf{{v}}$}}
\def\tvt{\mbox{$\boldsymbol{ \rho}^k$}}
\def\tvtt{\mbox{$\boldsymbol{ \rho}$}}
\def\tvto{\mbox{$\boldsymbol{ \rho}^k_{t_0}$}}
\def\tvtf{\mbox{$\boldsymbol{ \rho}^k_{t_1}$}}

\author{P. H\'EAS\thanks{INRIA Rennes \& IRMAR, Universit\'e de Beaulieu, 35042 Rennes, France ({\tt Patrick.Heas@inria.fr}) } \and   O. Hautecoeur\thanks{Exostaff GmbH, Riedstrasse 6, 64404 Bickenbach, Germany}   \and R. Borde\thanks{EUMETSAT, Eumetsat Allee, 64295 Darmstadt, Germany}  
   }

\begin{document}

\title{3D wind field profiles 
from hyperspectral  sounders: revisiting optic-flow from a meteorological  perspective}

\maketitle
\thispagestyle{empty}

\bigskip

\begin{abstract}
In this work, we present an efficient optic flow algorithm for the extraction of vertically resolved  3D atmospheric motion vector (AMV) fields  from incomplete hyperspectral image data measures by infrared sounders. 
The model at the heart of the energy to be minimized is consistent with atmospheric dynamics, incorporating  ingredients of thermodynamics, hydrostatic equilibrium and statistical turbulence.  
 Modern optimization techniques are deployed to design a low-complexity solver for the  energy minimization problem, which is non-convex, non-differentiable, high-dimensional and subject to physical constraints. In particular, taking advantage of the alternate direction  of multipliers  methods (ADMM), we show how to split the original high-dimensional problem into a recursion involving a set of standard and tractable optic-flow sub-problems. By comparing with the ground truth provided by the operational numerical simulation of the European Centre for Medium-Range Weather Forecasts (ECMWF), we show that the performance of the proposed method is superior to state-of-the-art optical flow algorithms in the context of real infrared atmospheric sounding interferometer (IASI) observations.

%A 3D AMV field, which is a function of the 3D spatial coordinates, is recovered from image sequences depicting temperature, specific humidity or ozone fields. The latter are measured at different pressure levels  by  infrared sounding.
%
%Based on thermodynamics, we design a cost function linking the image observations to the unknown 3D AMV field, and parametrized by a 3D diabatic heating field. We complement  this cost function by constraints originating from mass conservation and by state-of-the-art regularizers dedicated to  turbulent flows or sparse diabatic heating fields. 
%AMV reconstruction is then formulated as a constrained minimization problem. This problem has several specificities  which makes it particularly challenging.
%A first barrier  is the  high-dimensionality of the problem, which derives from the extra vertical dimension and the unknown diabetic heating field.
%A second obstacle is the non-differentiability and the non-convexity of some of the terms of the cost function: the non-differentiability stems from the use of the regularization terms  (e.g., to enforce sparsity) whereas the non-convexity originates from the temporal integration of thermodynamical laws.
%We propose an overall algorithmic framework to address this  problem. Our approach is based on  modern optimization techniques for non-differentiable/non-convex problems. As a consequence, we show that there exists a provably-convergent method estimating both the AMV field and the diabetic heating 3D distribution  with a complexity linear in the problem dimensions.

\end{abstract}

\begin{keywords}
3D atmospheric motion vector fields,  infrared atmospheric sounding interferometer, data assimilation, transport equation, vertical winds,  constrained optimization, wavelet-based optic flow.  \vspace{-0.cm}
\end{keywords}

%-------------------------------------------------------------------------
\section{Introduction}

Numerical weather prediction (NWP) models require the assimilation of meteorological observations.  NWP models must be continuously fed with a wide range of in situ observations, such as radiosondes, radars, buoys, aircraft measurements, and observations extracted from satellite data. The proportion of satellite data assimilated into numerical weather prediction models has increased considerably in recent years, as these data cover all regions of the Earth, especially the oceans and polar regions, where few in situ measurements are available~\cite{english2013impact}.  Atmospheric motion vectors (AMVs) derived from satellite images are the only wind observations with good global coverage~\cite{bedka2009comparisons}.  

Although AMVs have a significant positive impact on forecast scores, they provide information at only one level of the atmosphere and their retrieval is highly dependent on the presence of clouds. Therefore, they cannot meet the user requirements on  horizontal winds defined by the world meteorological organization in its report on applications of global numerical weather prediction:  horizontal and vertical resolutions of $50$ km and $1$ km, respectively~\cite{OSCAR}. 
On the other hand, there is currently no reliable alternative  based on satellite imagery to extract vertical wind profiles, and one must rely solely on conventional radiosonde measurements, wind profiles during ascent and descent of civil aircraft and by the  radar Doppler network. In contrast to satellite images, these observations are sparse and mainly collected in the northern hemisphere and over land areas. 
Therefore,  profiles of  horizontal and vertical  winds  have been identified as the most critical atmospheric variables that are not adequately measured by current or planned systems~\cite{WIGOS}. 

Infrared sounder measurements have been ranked as the second highest priority for horizontal wind profile retrieval, second only to  Doppler wind lidar  lidar measurements~\cite{OSCAR}. This ranking is evaluated as a mixture of several elements such as uncertainty, horizontal and vertical resolution, and frequency of recovered measurements, which are found to be primarily determined by the observing technique and the earth orbit (geostationary or low) of the satellite.
Several  works  have  been  conducted  to  investigate  the  extraction  of  AMV  profiles  from  moisture  and  temperature  fields  retrieved  from these hyperspectral  instruments~\cite{santek2019demonstration}. These studies have pointed out difficulties and limitations  linked  to  the  common  techniques  used  to  extract  AMVs. Most of the AMV extraction algorithms apply cross-correlation techniques on sequences of consecutive images \cite{bedka2005application}. The frequent lack of contrast in the  moisture  fields  retrieved  from  the hyperspectral  sounders  does  not  permit  the  unambiguous  identification  of  features in the images. Moreover,   the  moisture fields are characterized by large areas with missing observations making the tracking task even harder~\cite{apke2018relationships}.  This frequently leads to a poor matching, limiting overall AMV production. 

Energy minimization methods, known in the computer vision literature as optic flow algorithms, have shown promise as approaches in atmospheric science because of their good adaptation to the inherent physical nature of images, and because they can handle low contrast and missing observations, see~\cite{fortun2015optical} for a review on optic flow.   Among other meteorological studies, these techniques have been applied to the identification of flow boundaries~\cite{apke2020towards}, or adapted to the estimation of layered AMV fields at different altitudes in the troposphere by satellite measurements of cloud top temperature~\cite{heas2007layered}. Using the latter approach, the methodology has then been extended to the three-dimensional layered estimation of AMVs~\cite{heas2008three}. Finally, preliminary studies on the estimation of vertical profiles of dense 3D AMV fields using hyperspectral observations have been conducted in~\cite{borde2019winds}.

In line with the latter work, this paper proposes an improved optic flow methodology for the characterization of vertical profiles of 3D AMV fields processing noisy and incomplete hyperspectral satellite observations. The improvement is threefold. First, the model at the heart of the optic flow technique is based on a consistent physical modeling of atmospheric dynamics. Specifically, on the one hand, the data term incorporates a hydrostatic equilibrium constraint and a term penalizing deviations from the thermodynamic equation, providing a unified framework for the two well-known meteorological methods used for estimating vertical motion: the kinematic method and the adiabatic method~\cite{Holton92}.  On the other hand, the regularization term is adapted to statistical models of atmospheric turbulence~\cite{Heas11a,heas2012bayesian}. Secondly, the noise and the missing image observations are managed by the coupling of the estimation of AMV fields with the reconstruction of dense maps of temperature, humidity and ozone concentration. Lastly, modern optimization techniques are deployed to manage the optic flow problem taking the form of  a high-dimensional, non-convex and non-differentiable minimization problem subject to  physical constraints. More precisely, we  design alternate direction  of multipliers  methods (ADMM) to split the constrained high-dimensional problem into a set of unconstrained differentiable parallel minimizations~\cite{Boyd11}. These latter problems remain non-convex and high-dimensional. We rely on the wavelet expansion of the AMV fields~\cite{derian2013wavelets,Kadri13,Heas14,schmidt2019high,nicolas2023assessment} and employ large scale quasi-Newton methods~\cite{Nocedal99} to deal efficiently with   these problems.

This paper is organized as follows. In section~\ref{sec:model}, we present the atmospheric model on which our methodology is based. The optic flow estimation problem is then formalized in section~\ref{sec:ofProblem}, while a dedicated efficient solver is proposed in section~\ref{sec:5}. A numerical evaluation compares in section~\ref{sec:NumEval} the performance of the proposed method with a benchmark of state-of-the-art optic flow algorithms, for real observations of the Infrared Atmospheric Sounding Interferometer (IASI) . The ground truth used for the evaluation procedure is the synchronized data provided the operational numerical model of the European Centre for Medium-Range Weather Forecasts (ECMWF). Finally, a last section draws the conclusion.

\section{Geophysical Modeling}\label{sec:model}

We hereafter propose a set of constraints relating vertical profiles of  AMVs fields, which we will refer to 3D AMVs,   to the time evolution  of the spatial distribution  over the three dimensional space  of  three  meteorological quantities, namely temperature, specific humidity and ozone concentration.  The model relies on standard atmospheric dynamics  presented in~\cite{Holton92}.

\subsection{Pressure-Averaged Atmospheric States}

Let us  consider an isobaric coordinate system with the spatial coordinates denoted by
 $(s,p) \in \Omega \times \mathcal{Z} \subset \Rr^3$. The spatial distribution  at time $t \in \Rr$ of  the  three meteorological quantities of interest  is represented by the bounded multivariate function $\boldsymbol{\rho}(s,p,t)$ defined over $\Omega \times  \mathcal{Z} \times \Rr$ and taking its values in  $\Rr^3$.   
We are interested in relating the time evolution of function $\boldsymbol{\rho}(s,p,t)$ to the time evolution of  horizontal wind function   $\ww(s,p,t)$, both function being defined over $\Omega \times  \mathcal{Z} \times \Rr$ and taking their values respectively in  $\Rr^3$ and $\Rr^2$. To this aim, we consider the discretization of the pressure interval $\mathcal{Z}$  into a finite set  of decreasing pressure levels $\{p^k\}_{k=0}^K$  with $p^{k}>p^{k+1}$, yielding   pressure increments denoted by   $\delta p^k= p^{k}-p^{k+1}$ for  $k\in \mathcal{K}=\{0,\ldots,K-1\}$.
Using this vertical discretization, the  spatial distribution of  the  meteorological quantities over the  three dimensional space  is represented by a stack $\{\boldsymbol{\rho}^k\}_{k\in \mathcal{K}}$ of  pressure-averaged  functions $\boldsymbol{\rho}^k:\Omega \times \Rr \to \Rr^3$  defined as
\begin{eqnarray}\label{eq:rhoDef}
\boldsymbol{\rho}^k(s,t)&=&\frac{1}{\delta p^k}\int_{p^{k+1}}^{p^{k}} \boldsymbol{\rho}(s,p,t) dp,
\end{eqnarray}
and  a stack $\{\wwr\}_{k\in \mathcal{K}}$  of pressure-averaged horizontal winds $\wwr:\Omega \times \Rr \to \Rr^2$ as
%temperature fields~$T^k$, specific humidity fields $q^k$, diabetic heating fields $h^k$ and horizontal wind fields~$\mathbf{v}^k$. For $x$ standing either for variable $T$, $q$, $\oo$, $h$,  $\mathbf{v}$, $u$ or $v$, these  pressure-averaged quantities are defined as:
\begin{eqnarray}\label{eq:vDef}
\wwr(s,t)&=&\frac{1}{\delta p^k}\int_{p^{k+1}}^{p^{k}} \ww(s,p,t) dp.
\end{eqnarray}
In addition,  we introduce  the vertical wind functions $\omega(s,p,t)$ defined over $\Omega \times  \mathcal{Z} \times \Rr$ and taking its values in  $\Rr$. The vertical winds taken at the discrete set of pressure levels yield the stack  $\{\omega^k(s,t)\}_{k=0}^K$ whose elements are 
$
\omega^k(s,t)=\omega(s,p^{k},t).
$

\subsection{Time-Integrated Atmospheric Dynamics}

As detailed in Appendix~\ref{sec:appGeophys}, simplified geophysical models describing the time evolution of  temperature and specific humidity  may be obtained  by neglecting diabetic heating in the first law of thermodynamics. Moreover, similar simplified model may be obtained making the common assumption that ozone concentration evolves  as a passive scalar transported by the flow. These  dynamical models take  the form of the transport equation described hereafter. Assuming some mild smoothness condition\footnote{ The horizontal and vertical wind fields are asumed to be $\wwr(s,t)\in  ( \mathcal{C}^1(\Omega \times \mathbb{R}))^2$, $\omega^{k}(.,t)\in  \mathcal{C}^1(\Omega \times \mathbb{R})$  and  Lipschitz continuous. },  the pair $(\xx_{t_0}^{k},\xx_{t_1}^{k})$ can be identified to the  solution $\tvt(s,t)$ taken on the points of the grid $\Omega_m$, at times $t_0$ and $t_1$ (with $t_0 < t_1$) of   the transport  equation of  initial condition $\tvto(s)$
\begin{equation}\label{eq:OFC}
 \left\{\begin{aligned}
&\frac{\partial \tvt}{\partial t}(s,t)  + \wwr(s,t) \cdot\nabla_s  \tvt(s,t) +\frac{1}{2 }(\gamma^k\omega^{k}(s,t)+\gamma^{k+1}\omega^{k+1}(s,t))=0,\\
&\tvt(s,t_0)=\tvto(s)
\end{aligned}\right.,
\end{equation}
where $ \wwr(s,t) \cdot\nabla_s  \tvt(s,t) $ denotes the scalar product of the gradients of each of the three components of $ \tvt(s,t) $ with $\wwr(s,t)$, and where the $\gamma^k\in \Rr^3$  are known physical constants. It is well known   that under mild conditions, when $\delta t\triangleq t_1-t_0$ is a small increment, we obtain from \eqref{eq:OFC} by time integration the Lagrangian form of the dynamics which we will call the {\it warping  constraints}\footnote{
Let the function $t \rightarrow {\bf X}_{ t_0}^t(s)$ be the characteristic curves of the partial differential equation~\eqref{eq:OFC}, solution of the system:
 \begin{equation*}
 \left\{\begin{aligned}
&\frac{d }{d t}{{\bf X}}_{ t_0}^t(s)  = \wwr({{\bf X}}_{ t_0}^t(s),t) \\
&{{\bf X}}_{ t_0}^{t_0}(s)=s
\end{aligned}\right..
\end{equation*}
It follows from \eqref{eq:OFC} that the field at initial time  satisfies~\cite{raviart1983introduction} $$\tvto(s) =\tvt({\bf X}^{t_1}_{t_0}(s),t_1) +\int_{t_0}^{t_1} \frac{1}{2 }(\gamma^k\omega^{k}({{\bf X}}_{ t_0}^{t'}(s),t')+\gamma^{k+1}\omega^{k+1}({{\bf X}}_{ t_0}^{t'}(s),t')) dt'.$$
We then  obtain \eqref{eq:warpingDFD} by assuming that  $\int_{t_0}^{t_1}\wwr({{\bf X}}_{ t_0}^{t'}(s),t')dt'=\delta t\wwr(s,t_0)$, and that the vertical winds $\omega^{k}$ and $\omega^{k+1}$ are constant in the time interval $[t_0,t_1]$.
% $\omega^{k}({{\bf X}}_{ t_0}^{t'}(s),t')= \omega^{k}(s,t_0)$ and $\omega^{k+1}({{\bf X}}_{ t_0}^{t'}(s),t'))=\omega^{k+1}(s,t_0))$ for $t'\in [t_0,t_1].$
}
    
 \begin{align}\label{eq:warpingDFD}
 \tvto(s) =\tvtf(s+  \delta t\, \wwr(s,t_0))+\frac{\delta t}{2 }(\gamma^k\omega^{k}(s,t_0)+\gamma^{k+1}\omega^{k+1}(s,t_0)).
 \end{align}
 
 \subsection{Constraint on Hydrostatic Equilibrium}
 
Besides, mass conservation provides in addition to~\eqref{eq:warpingDFD} a geophysical constraint on the structure of horizontal and vertical winds. Indeed, in isobaric coordinates mass conservation writes:
\begin{equation}\label{ICE_iso}
-\frac{\partial \omega(s,p,t)}{\partial p}= \nabla_s \cdot \ww(s,p,t).%  \left( \frac{\partial u}{\partial x}+\frac{\partial v}{\partial y} \right)_p.
\end{equation}
Vertical integration of  \eqref{ICE_iso} in
the pressure interval $[p^{k},p^{k+1}]$ yields\footnote{under the assumption on the vertical wind function mentioned earlier.}  at time $t_0$ for $k \in \mathcal{K}$ what we will call the {\it hydrostatic constraints}
\begin{align}\label{ICE}
\omega^{k}(s,t_0)-\omega^{k+1}(s,t_0) = \delta p^k\nabla_s \cdot \mathbf{v}^k(s,t_0),
\end{align}
 with the boundary conditions   $\omega^{0}(s,t_0)=0$ and $\omega^{K}(s,t_0)=0$ for any $s\in \Omega$. These boundary conditions on vertical winds can be argued as follows: 
 for the lowest layer ($k$=0), the Earth boundary condition
implies zero vertical winds, while for the highest
layer ($k$=K) a reasonable assumption is that vertical wind can be
neglected at the tropopause which acts like a
cover.

\subsection{Spatial Discretization}
We now describe the spatial discretization of the continuous functions introduced in the previous sections. Consider the image grid  $$\Omega_m=\{s\in \Omega : s=\varkappa(j), j=1,\ldots,m\},$$
where   $\varkappa(j)$  is the function returning the spatial position corresponding to index $j$.   Using this spatial discretization of the bi-dimensional domain $\Omega$, we finally define at time $t$ the stack $\xx_{t}^\star$ of images in $\Rr^{3Km}$, whose $k$-th layer is a vector with $m$ tri-variate components  
$\xx^k_{t}(j)=\boldsymbol{\rho}^k(\varkappa(j),t), \quad \forall j : \varkappa(j)\in \Omega_m. $
We are interested in the pair of stack of pressure-averaged images at time $t_0$ and $t_1$, \ie, $(\xx_{t_0}^\star,\xx_{t_1}^\star)\in (\Rr^{3Km})^2$. Related to the image grid $\Omega_m$, we  also define the AMVs: a stack of pressure-averaged  horizontal displacement fields $\mathbf{d}^\star\in \Rr^{2Km}$, whose $k$-th layer is a vector of $m$ bi-variate components defined as
$\mathbf{d}^k(j)=\delta t\, \wwr(\varkappa(j),t_0), \quad \forall j : \varkappa(j)\in \Omega_m,$
 and a stack of vertical displacement fields $\boldsymbol{\omega}^\star \in \Rr^{(K+1)m}$ located at the frontiers of the layers, whose $k$-th component is the vector  with $m$  components defined as 
 $\boldsymbol{\omega}^k(j)=\omega^k(\varkappa(j),t_0), \quad \forall j : \varkappa(j)\in \Omega_m$,  where according to the boundary conditions, we have  for any $j$ the  conditions   $\boldsymbol{\omega}^0(j)=0$ and $\boldsymbol{\omega}^K(j)=0$.

\subsection{3D AMV Model}
We are now ready to define the 3D AMV model, which relies on pressure-averaged time-integrated and spatially-discretized  physical conservation laws.

The first constraint defining our discrete model for 3D AMVs relies on  mass conservation given in its pressure-averaged and spatially-discretized form by the hydrostatic constraints~\eqref{ICE}. Applied on the pixel grid, the latter takes  for $k\in \mathcal{K}$  the form of  
\begin{align}\label{ICE2} 
\boldsymbol{\omega}^k-\boldsymbol{\omega}^{k+1} = \delta p^k \textrm{div} (\mathbf{d}^k).
\end{align}
Let us note that the components of the vector $\textrm{div} (\mathbf{d}^k)$ in $\Rr^m$ is  the divergence  at location in $\Omega_m$ of a continuous representation of $\mathbf{d}^k$ on the bi-dimensional domain $\Omega$.

Then, in order to  complement  this mass  constraint,  for each  couple $(\xx_{t_0}^{k},\xx_{t_1}^{k})$ we will need to assume a continuous model for interpolating the images $\xx_{t_1}^{k}$ outside of $\Omega_m$. Taking the warping constraints~\eqref{eq:warpingDFD} at points in $\Omega_m$, we rewrite  the warping model  for the $k$-th layer  as
\begin{align}\label{eq:warpingModel}
\mathbf{x}_{t_0}^k=\mathcal{W}(\xx_{t_1}^k, \mathbf{d}^k,\boldsymbol{\omega}^k,\boldsymbol{\omega}^{k+1}),
\end{align}
where operator $\mathcal{W}: \Rr^{3m} \times \Rr^{2m}\times \Rr^{m}\times \Rr^{m}\rightarrow \Rr^{3m} $  in \eqref{eq:warpingModel} warps the stack of images $\xx^k_{t_1}$ into  $\xx^k_{t_0}$
according to the  AMV pressure-averaged  horizontal component $\mathbf{d}^k$ and to the vertical winds $\boldsymbol{\omega}^k$ and $\boldsymbol{\omega}^{k+1}$ on the upper and lower layer boundary. 
 The $j$-th component output $\mathcal{W}_{j}: \Rr^{3m} \times \Rr^{2m} \times \Rr^{m}\times \Rr^{m} \to \Rr^3$ of operator $\mathcal{W}$ is the 
 function   defined as  $\mathcal{W}_{j}(\xx_{t_1}^k, \mathbf{d}^k,\boldsymbol{\omega}^k,\boldsymbol{\omega}^{k+1}) =$
\begin{align}\label{eq:splineRepr}
&\sum_{i\in \mathcal{V}(\varkappa(j)+\mathbf{d}^k(j))} \xx_{t_1}^k(i)\varphi_i(\varkappa(j)+\mathbf{d}^k(j))  
-\frac{\delta t}{2 }(\gamma^k\boldsymbol{\omega}^k(j)+\gamma^{k+1}\boldsymbol{\omega}^{k+1}(j)),
 \end{align}
where $\mathcal{V}(\varkappa(j)+\mathbf{d}^k(j))$ denotes a subset of  indices corresponding to the ``neighborhood'' of point $\varkappa(j)+\mathbf{d}^k(j)$. The family $\{\varphi_i\}_{i=1}^m$ with  $\varphi_i: \Omega_m \to \Rr $ is chosen to be the bi-dimensional cubic cardinal   splines  interpolation functions  \cite{Unser91}. 
Note that the stack of images $\xx_{t_0}^k$ is a deterministic function of $\xx_{t_1}^k$ and $\mathbf{d}^k$, $\boldsymbol{\omega}^k$ and $\boldsymbol{\omega}^{k+1}$.  We remark that $\mathcal{W}$ is linear in its first, third and fourth arguments and  non-linear in its second one as long as $\varphi_i$'s are  non-linear. 

To summarize, the 3D AMV model is  \eqref{ICE2}--\eqref{eq:splineRepr}.

\section{ Formulation of the Estimation Problem}\label{sec:ofProblem}
In this section, we expose our methodology to estimate pressure-averaged  vertical profiles of 3D AMV fields. The estimation relies on vertically resolved and partially observed fields of temperature, specific humidity and ozone. 
\subsection{Partial Observations}
We refer to   the  stack  $(\mathbf{d}^\star,\boldsymbol{\omega}^\star,\xx_{t_1}^\star)$ of $K$ triple  satisfying the 3D AMV model  \eqref{ICE2}--\eqref{eq:splineRepr}, as the ``ground truth''. The ground truth related to  the $k$-th layer of the stack will be denoted by the triple  $(\mathbf{d}^{k,\star},{{\boldsymbol{\omega}^k}^\star},\xx_{t_1}^{k,\star})$. Of course the  ground truth is usually unknown, but we may assume some generative model relating the ground truth to the available noisy and partial observations. More explicitly,  let   $ \Omega^{t_0,k}_{obs},\,  \Omega^{t_1,k}_{obs} \subseteq \Omega_m$ denote the set of spatial locations of the image grid  related to  the observed components of $\xx_{t_0}^{k,\star}$ and $\xx_{t_1}^{k,\star}$. 
 The set of observations is  $$\yy=\{\yy_{t}^{k}({j}) \in \Rr^3 : \varkappa(j)  \in  \Omega^{t,k}_{obs} ; t \in \{t_0,t_1\};\,k \in \mathcal{K}\} \in \mathcal{Y},$$
where  $ \mathcal{Y} \in \Rr^{3Km}$ and the three components of $\yy_{t}^{k}({j})$  are those of  $\xx^{k,\star}_t(j)$ up to some additional centered Gaussian noise of covariance $\mathbf{I}_{\sigma_{obs}^2}=\diag(\sigma_{obs}^{2},\sigma_{obs}^{2},\sigma_{obs}^{2})$. Therefore, according to the warping model \eqref{eq:warpingModel}, the observations at the $k$-th layer are generated from the ground truth as
\begin{align*}
\yy_{t_0}^{k}({j}) &=\mathcal{W}_j(\xx^{k,\star}_{t_1}, \mathbf{d}^{k,\star},{{\boldsymbol{\omega}^k}^\star},\boldsymbol{\omega}^{k+1,\star}) + \mathcal{N}(0,\mathbf{I}_{\sigma_{obs}^2}),\quad \varkappa(j)  \in \Omega^{t_0,k}_{obs}, \\
\yy_{t_1}^{k}({j}) &=\xx^{k,\star}_{t_1}(j) + \mathcal{N}(0,\mathbf{I}_{\sigma_{obs}^2}),\quad \varkappa(j)  \in \Omega^{t_1,k}_{obs}.
 \end{align*}
 More involved schemes use alternative noise assumptions to take into account noise correlation and non-quadratic deviations, see references in~\cite{Butler:ECCV:2012}. 

%\mathcal{P}_T(\boldsymbol{T}^k_{t_1}, \mathbf{d},\boldsymbol{h}^k,\boldsymbol{\omega}^k,\boldsymbol{\omega}^{k+1})= \sum_{\mathbf{s}\in \Omega}  &|T^k(\mathbf{s}+   \mathbf{d}(\mathbf{s}),t+1)-T^k(\mathbf{s},t) \\
% &+\frac{h^k(\mathbf{s})}{c_p}  -\frac{\delta t}{2} (\gamma^k(\mathbf{s})\omega^{k}(\mathbf{s})+\gamma^{k+1}(\mathbf{s})\omega^{k+1}(\mathbf{s}))  |^2,\nonumber\\
%\end{align*}

\subsection{Optimization Problem}\label{sec:3}  \label{sec:formulationOptim}
 Given the incomplete observations $\yy$ of temperature, humidity and ozone, the problem is  the estimation of    3D AMVs, \ie,  $(\mathbf{d}^\star,\boldsymbol{\omega}^\star) \in \Rr^{K2m} \times  \Rr^{(K-1)m}$,  together with the estimation of the stack of  images $\xx_{t_1}^\star \in \Rr^{3Km}$. The vector  parametrizing the 3D AMV model \eqref{ICE2}--\eqref{eq:splineRepr} is thus  $\boldsymbol{\theta}^\star=({\mathbf{d}^\star}^\intercal,{\boldsymbol{\omega}^\star}^\intercal,{\xx_{t_1}^\star}^\intercal)^\intercal \in \Rr^n$, with $$n=(6K-1)m.$$ Let us point out that the dimension $n$ of the 3D AMV model is typically huge: considering the realistic scenario of $K=2^4$ layers and images of size $m=2^{9} \times 2^{9}$, the number of degrees of freedom $n$ of the 3D AMV model is greater than  24 millions. %humidity maps $\{q_{obs}^k(\mathbf{s},t)\}_{k=0,\mathbf{s} \in \Omega_{obs}}^{K-1} $  and ozone maps $\{o_{obs}^k(\mathbf{s},t)\}_{k=0,\mathbf{s} \in \Omega_{obs}}^{K-1} $ at time coordinate $t$ and $t+1$. \\
Let us consider an optimization variable $\boldsymbol{\theta}=(\mathbf{d}^\intercal,\boldsymbol{\omega}^\intercal,\xx_{t_1}^\intercal)^\intercal \in \Rr^n$  of dimension identical to  $\boldsymbol{\theta}^\star$.   Our approach is based on the resolution  of the  {\it hard constrained}   optimization problem   
\begin{equation}\label{eq:probDef}
\left\{\begin{aligned}
 & \argmin_{(\mathbf{d},\boldsymbol{\omega},\xx_{t_1})}  \mathcal{J}{(\mathbf{d},\boldsymbol{\omega},\xx_{t_1},\yy)} + \boldsymbol{\alpha}_{\mathbf{d}} \mathcal{R}_d(\mathbf{d})+  \boldsymbol{\alpha}_{\xx}\mathcal{R}_\xx(\xx_{t_1}),\\
 &\textrm{s.t.}\quad h(\mathbf{d}, \boldsymbol{\omega})=0,
  \end{aligned}\right.
\end{equation}
where the cost $\mathcal{J}$ denotes  the data term,  $ \mathcal{R}_d$ and $ \mathcal{R}_\xx$  denote some regularizers, and $h$ denotes some constraint. The components of parameter vectors $\boldsymbol{\alpha}_d=(\alpha_d^1,\cdots,\alpha_d^K)^\intercal $ and  $\boldsymbol{\alpha}_{\xx}=(\alpha_\xx^1,\cdots,\alpha_\xx^K)^\intercal $ are positive reals, which we assumed pre-defined by  expert knowledge. 

Alternatively, substituting the { hard} constraint  in  \eqref{eq:probDef} by a  quadratic penalization, we may also consider the {\it soft constrained} minimization problem   
\begin{equation}\label{eq:probDefSoft}
\begin{aligned}
 & \argmin_{(\mathbf{d},\boldsymbol{\omega},\xx_{t_1})}  \mathcal{J}{(\mathbf{d},\boldsymbol{\omega},\xx_{t_1},\yy)} + \boldsymbol{\alpha}_{d} \mathcal{R}_d(\mathbf{d})+  \boldsymbol{\alpha}_{\xx}\mathcal{R}_\xx(\xx_{t_1})+\rho h(\mathbf{d}, \boldsymbol{\omega})^2,\\
  \end{aligned}
\end{equation}
where $\rho $ is a given positive real. 
Let us provide some precisions on the components of problem \eqref{eq:probDef} and \eqref{eq:probDefSoft}. 

\subsubsection{The Linear Constraint}
The  constraint  in \eqref{eq:probDef} or the  function in \eqref{eq:probDefSoft} is linear and defined as
\begin{equation}\label{eq:linearConst}
 h(\mathbf{d}, \boldsymbol{\omega})=\mathbf{D} \mathbf{d}-\mathbf{L}  \boldsymbol{\omega}
\end{equation}
with matrices $\mathbf{L} \in  \Rr^{Km \times (K-1)m}$ and $\mathbf{D} \in \Rr^{Km \times 2Km}$. The hydrostatic balance   \eqref{ICE2} under the vertical wind boundary conditions can be rewritten in the form of the matrix-vector products  \eqref{eq:linearConst} by choosing specific matrices $\mathbf{L}$ and $\mathbf{D}$. 
%taking the form for $ 1\le k \le K-1$ and $s\in \{1, \cdots, n\}$ of 
%\begin{align}\label{eq:constDivW}
%\boldsymbol{\omega}^{k}(s)-\boldsymbol{\omega}^{k+1}(s) =\delta p^k \textrm{div}\,\mathbf{d}^k\,(s),
%\end{align}
%where operator $\textrm{div}:(\Rr^n)^2 \to \Rr^n$ is a discrete approximation of the divergence operator for the finite grid $\Omega_n$, and where we have used the assumption that the horizontal winds  $\mathbf{v}^k(\varkappa(s))=\mathbf{d}^k(s)$, so as the vertical winds ${\omega}^k(\varkappa(s),t)$ and ${\omega}^k(\varkappa(s),t+1)$, are constant  within the time interval $[t, t+1]$
We remark that in this case this linear constraint  generates an overdetermined system in $ \boldsymbol{\omega}$:  there are $(K-1)$  unknown $m$-dimensional  vertical wind fields,   for $K$ $m$-dimensional constraints. However,  $m$-linear rows of $\mathbf{L}$ are linearly dependent. As a consequence, there exists a unique $\boldsymbol{\omega}$ satisfying the linear constraint, for  any value of $\mathbf{D}\,{\mathbf{d}}$. Therefore in problem ~\eqref{eq:probDef}, $\boldsymbol{\omega}$ can be expressed as a deterministic function of ${\mathbf{d}}$:  $\boldsymbol{\omega}=\mathbf{L}^{\dagger}\mathbf{D}\,{\mathbf{d}}$. As we shall see, optimization with the linear  constraint $\mathbf{L}\, \boldsymbol{\omega}=\mathbf{D}\,{\mathbf{d}}$  will be preferable  in terms of algorithmic implementation to the use of the deterministic function $\boldsymbol{\omega}={\mathbf{L}^{\dagger}\mathbf{D}\,{\mathbf{d}}} $, as it offers the possibility to parallelize the optimization procedure.   

\subsubsection{The Data Term}
The data-term  $\mathcal{J}$ in  \eqref{eq:probDef}  is    a  function  depending on the  vector $\boldsymbol{\theta}=(\mathbf{d}^\intercal,\boldsymbol{\omega}^\intercal,\xx_{t_1}^\intercal)^\intercal \in \Rr^n$ and on observations $\yy \in  \mathcal{Y}$. Its aim is to penalize discrepencies between  given states $\xx_{t}$  at time $t=t_0$ and $t_1$ and the related  observations (when available), where we  recall that $\xx_{t_0}$ is a deterministic mapping of $\xx_{t_1}$ by  the  warping model \eqref{eq:warpingModel}.
More precisely, we define the residual  function $\boldsymbol{\delta}_{}: \Rr^n \times \mathcal{Y} \to \Rr^{6m}$ such that $\boldsymbol{\delta}_{}(\boldsymbol{\theta},\yy)=\begin{pmatrix}{\boldsymbol{\delta}_{t_0}}(\boldsymbol{\theta},\yy)\\ {\boldsymbol{\delta}_{t_1}}(\boldsymbol{\theta},\yy)\end{pmatrix}$, with  the $(k,j)$-th component of ${\boldsymbol{\delta}_{t}}(\boldsymbol{\theta},\yy)$   defined for $k\in\mathcal{K}$ and $\varkappa(j) \in \Omega_m$
 as: 
 \begin{align}\label{eq:defDeltax}
{\boldsymbol{\delta}_{t}^{k,s}}(\boldsymbol{\theta},\yy)=
\left\{\begin{aligned}
&\xx^{k}_t({j})-\yy^{k}_t({j}) \quad \textrm{if}\quad \varkappa(j) \in \Omega^{t,k}_{obs}\\
&0  \quad \hspace{2.3cm}\textrm{else}
 \end{aligned}\right. .
\end{align}
We have assumed that the observation noise is  Gaussian and uncorrelated. Under this Gaussian assumption, the most likely state is the one minimizing  the square of the residual function norm. We thus define the data-term simply as     
% More precisely, we assume that the so-called data-term DFD functional
%\begin{equation}\label{objFunction}
%\delta y( \mathbf{d})\triangleq\frac{1}{2}\|y_{1}(\xx+\mathbf{d}(\xx))-y_0(\xx)\|^2; \hspace{0.25cm} \mathbf{d} \in L_{div}^2([0,1]^2),\; y_0, y_1  \in L^2([0,1]^2),
% \end{equation} 
\begin{align}\label{eq:likelyGauss}
\mathcal{J}(\boldsymbol{\theta},\yy)=\frac{1}{2} \|\boldsymbol{\delta}(\boldsymbol{\theta},\yy)\|^2_2.
\end{align}

\subsubsection{The Regularizers}
It is well known that solving problem \eqref{eq:probDef} or \eqref{eq:probDefSoft} with no regularization, \ie,  setting the regularization parameters to zero, yields a severely ill-conditioned problem. In the present meteorological context, some regularizers are more appropriate than others to the specificity of the geophysical flows.

Indeed, many options for $\mathcal{R}_d$ have been proposed in the computer vision literature, starting with the famous first-order optic flow regularization~\cite{Horn81}.  However, this model is not suitable for fluid flows, as it smoothes out the vortex and divergence structures. Second-order schemes are more relevant for fluid flows. Among others, popular approaches are to smooth the gradient of the divergence and vorticity~\cite{Suter94} or the higher order derivatives~\cite{Kadri13}. Recent  schemes~\cite{Tafti11,Heas14}, 
 propose a quadratic regularization   taking the form of
\begin{align}\label{eq:DefR2}
\boldsymbol{\alpha}_d\mathcal{R}_d(\mathbf{d})= \frac{1}{2}\sum_{k=0}^{K-1}{\alpha_d^k} \|  {\boldsymbol{\Delta}_m}\mathbf{d}^k\| ^2_2.
\end{align}
%\begin{align}\label{eq:DefRGen}
%\boldsymbol{\alpha}_d\mathcal{R}_d(\mathbf{v})&= \frac{1}{2}\|\mathbf{G}_H^\intercal \mathbf{d}\|^2_2.
%%&= \sum_{k=0}^{K-1} \sum_{\mathbf{s} \in \Omega}  \sum_{\ell \in \nu_\mathbf{s} }({\boldsymbol{\Delta}_m^{H}}^\intercal _\ell \mathbf{d}^k)^2,
%\end{align}
where the operator ${\boldsymbol{\Delta}_m} : \Rr^{2m} \to \Rr^m$  
 is defined as a finite approximation of the bi-dimensional Laplacian operator  applied  to the displacement field ${\mathbf{d}^k}$. The regularization \eqref{eq:DefR2} provides a relevant prior structure for geophysical flows, which characterizes an isotropic  self-similar structure for 2D  turbulence~\cite{heas2012bayesian,Heas14}.
 
With respect to the $\mathcal{R}_\xx$ regularizer, a standard approach to reconstructing dense fields from the observation of incomplete maps is to rely on sparse priors~\cite{foucart2013mathematical}.  In particular, the geophysical variables of temperature, humidity, and ozone concentration, which are solutions of a transport equation of the form ~\eqref{eq:OFC}, admit a sparse decomposition in some well-chosen basis as argued in~\cite{candes2005curvelet}.  Specifically, we will assume that the temperature, humidity, and ozone concentration images in the $\xx_{t_1}$ stack are sparse in a dictionary of two-dimensional interpolation functions. 
To this aim,  we  represent  the components of the state ${\xx}^k_{t_1}\in \Rr^{3m}$ using three series. Let ${\xx}^{k,\ell}_{t_1} \in \Rr^m$  for $\ell=1,2,3$ be the $\ell$-th image  of the stack  ${\xx}^{k}_{t_1}$. We define  
% $\mathcal{P}_{\mathbf{d}_{t}}$ admits the following series representation:
$\mathbf{x}^{k,\ell}_{t_1}=\mathbf{\bar \Phi} \,\mathbf{c}^{k,\ell},$
 where   $\mathbf{\bar \Phi}\in \Rr^{m \times m}$ is an orthonormal basis   $\{\phi_i\}_{i=1}^m$ with $\phi_i:\Omega_m \to \Rr$, and   with  the coefficient vector $\mathbf{c}^{k,\ell}\in \Rr^{m}$.   Denoting by  $\mathbf{c}^k\in \Rr^{3m}$ the concatenation  of the  $\mathbf{c}^{k,\ell}$'s for $\ell=1,2,3$,  and furthermore denoting by  $\mathbf{c}\in \Rr^{3Km}$ the concatenation  of the  $\mathbf{c}^{k}$'s for $k\in \mathcal{K}$, we define  operator  $\boldsymbol{\Phi}: \Rr^{3Km} \to \Rr^{3Km} $   such that   
$
\mathbf{x}_{t_1}=\mathbf{ \Phi} \,\mathbf{c}.
%&=\mathbf{D}\mathbf{c}^k_{\xx},
$ Coefficients $\mathbf{c}^{k}$ are related to the image stack  $\xx^{k}_{t_1}$ through the latter decomposition.
  It is now well established that the  sparse assumption may be modeled (under some specific assumption) by  an $\ell_1$ norm penalization~\cite{foucart2013mathematical}, \ie, 
 \begin{align}\label{eq:regC}
 \boldsymbol{\alpha}_{\xx}\mathcal{R}_\xx(\xx_{t_1})= \frac{1}{2}\sum_{k=0}^{K-1} {\alpha}_{\xx}^k \|\mathbf{c}^{k}\|_1.%Sparse models such as \eqref{eq:sparseprior} have been considered with success in many contributions dealing with inverse problems. 
 \end{align}

%We finally notice that linear  constraints  \eqref{eq:constDivW} represent the balance between pressure-averaged horizontal displacement divergence and vertical winds at the boundaries. Therefore, through this equilibrium, the regularity of the divergence field of horizontal displacements enforced by $\mathcal{R}_d$ will  impact the regularity of  vertical wind fields. %Constraints   \eqref{eq:constDivW}  thus substitutes regularization on $\boldsymbol{\omega}$. 

  \section{Efficient Solvers}\label{sec:5}

As  mentioned earlier, the number of variables involved in the constrained minimization problem \eqref{eq:probDef} is typically huge. Moreover, the objective function  is non-convex (due to the data term $\mathcal{J}$) and  non-differentiable (due to the regularizer $\mathcal{R}_\xx$). 
Accessing the minimum of the constrained minimization problem in this context is obviously a difficult task. Although the convergence to a global minimum is usually out of reach for any deterministic optimization procedure,  local minima can still constitute relevant  approximations.  Nevertheless, only specifically-dedicated procedures  can address efficiently such a high-dimensional, non-convex, non-differentiable and constrained optimization problem.  

In the sequel, we propose  optimization methods based on the  \textit{alternating direction method of multipliers} (ADMM)~\cite{Boyd11}.  We will also specify some convergence issues for the ADMM procedure particularized to   our  problem.
 ADMM  appeared more than ten years ago in the optimization community to deal with large-scale constrained and non-differentiable optimization problems. This type of method is known for its robustness (convergence to a local minimum is
is ensured under very mild conditions) and for its fast convergence to an acceptable accuracy (typically, a few tens of iterations are sufficient). We refer the reader to Appendix~\ref{app:A} for a brief description of the ADMM framework.

\subsection{Dealing with  Constraints and Non-Differentiability}\label{sec:DealingNonDiff}
In order to solve a non-differentiable constrained optimization problem, ADMM transforms it into a three-stage recursion in which the first two stages are standard unconstrained minimization problems.  To derive ADMM recursions, we reformulate the constrained optimization problem \eqref{eq:probDef}.  To this aim,   we  add a new variables to the problem:  $\boldsymbol{\tilde c}$ made of the concatenation of  vectors $\boldsymbol{\tilde c}^0,...,\boldsymbol{\tilde c}^{K-1}$, which are counterbalanced by the inclusion of a new constraint $\boldsymbol{\tilde c}=\boldsymbol{ c}$. 
Problem~\eqref{eq:probDef}
 can then be reshaped as
 \begin{equation}\label{eq:ADMMProb}
\left\{\begin{aligned}
%&(\mathbf{e}_{\boldsymbol{  d}} , \boldsymbol{\omega},\boldsymbol{\tilde \omega},\boldsymbol{h}, \boldsymbol{\tilde h},\boldsymbol{c},\boldsymbol{\tilde c},\boldsymbol{\epsilon}) 
%=\\
& \argmin_{\mathbf{d},\boldsymbol{\omega},\boldsymbol{c},\boldsymbol{\tilde c}} \mathcal{J}(\mathbf{d}, \boldsymbol{\omega},\boldsymbol{c})
+ \boldsymbol{\alpha}_{d} \mathcal{R}_d(\mathbf{d})  + \boldsymbol{\alpha}_{\xx}\mathcal{R}_\xx(\mathbf{ \Phi} \boldsymbol{\tilde c}),\\
&\textrm{s.t.}\quad \begin{pmatrix} \mathbf{D}&- \mathbf{L}&\mathbf{0}\\  \mathbf{0} & \mathbf{0} &\mathbf{I} \end{pmatrix}\begin{pmatrix}{\mathbf{d}}\\ \boldsymbol{\omega}    \\\boldsymbol{c}\end{pmatrix} +
 \begin{pmatrix}   \mathbf{0}\\  - \mathbf{I}\end{pmatrix} \boldsymbol{\tilde c}  =
 \begin{pmatrix} \mathbf{0}\\ \mathbf{0} \end{pmatrix} .
  \end{aligned}
  \right.
\end{equation}
As detailed in Appendix \ref{app:B}, this  constrained optimization problem fits the generic ADMM  procedure exposed in appendix \ref{app:A}. It yields a  solution  obtained by iterating until convergence  
 the three following  steps: 
\begin{align}\label{eq:step1_ADMM}
&({\mathbf{d}}^{(i+1)},\boldsymbol{\omega}^{(i+1)} ,\boldsymbol{c}^{(i+1)}) =\argmin_{{\boldsymbol{d}},\boldsymbol{\omega},\boldsymbol{c}}\,  \mathcal{J}(\mathbf{d}, \boldsymbol{\omega}, \boldsymbol{c}) + \boldsymbol{\alpha_d}\mathcal{R}_d(\mathbf{e}_{\boldsymbol{d}} )  + \frac{\rho}{2} \|  \mathbf{c} -   \mathbf{\tilde c} \,^{(i)}+  \mathbf{u}_{\mathbf{c}}\,^{(i)} \|^2_2 \nonumber \\
%& \hspace{3cm} +\gamma\, \mathcal{C}(\boldsymbol{\omega}-  \boldsymbol{\omega} ^{(j)},\boldsymbol{h}-  \boldsymbol{h} ^{(j)},\boldsymbol{q}-  \boldsymbol{q} ^{(j)},\boldsymbol{T}-  \boldsymbol{T} ^{(j)},\boldsymbol{o}-  \boldsymbol{o} ^{(j)})  \\
&\hspace{5cm} +\frac{\rho}{2}\| \mathbf{D}\mathbf{d}-\mathbf{L}\,\boldsymbol{\omega} + \mathbf{u}_{{\mathbf{d}}}^{(i)} \|^2_2   , \\
 &{\mathbf{\tilde c}^{k,\ell}\,}^{(i+1)}(s)=\mathrm{soft}_{\frac{{\alpha}^k_{\xx}}{\rho}} \left(\mathbf{c}^{k,\ell}\,^{(i+1)}(s) + \mathbf{u}^{k,\ell}_{\boldsymbol{c}}\,^{(i)}(s)\right),
  k\in \mathcal{K}, \ell\in\{1,2,3\},  s:\chi(s)\in \Omega_m ,\label{eq:step2_ADMM_prox}\\
  &\left\{
 \begin{array}{ll}
 \mathbf{u}_{{\mathbf{d}}}^{(i+1)}=& \mathbf{u}_{{\mathbf{d}}}^{(i)} + \mathbf{D}\,\mathbf{d}^{(i+1)}-\mathbf{L}\, \boldsymbol{\omega}^{(i+1)}\\
%{\mathbf{u}_{{\mathbf{d}^k}}}^{(i+1)}=& {\mathbf{u}_{{\mathbf{d}^k}}}^{(i)} + \delta p^k\textrm{div}{\mathbf{d}^k}^{(i+1)}-{\boldsymbol{\omega}^{k}}^{(i+1)}+{\boldsymbol{\omega}^{k+1}}^{(i+1)}, k\in \mathcal{K} \\
\mathbf{u}_{\boldsymbol{c}}^{(i+1)}\,\,\,\,\,\, =& \mathbf{u}_{\boldsymbol{c}}^{(i)} + {\boldsymbol{c}}^{(i+1)} -{\boldsymbol{\tilde c}}^{(i+1)}
\end{array}
 \right.
   \label{eq:step3_ADMM_prox},
\end{align}
where $\rho>0$ and where
\begin{align} \label{eq:softthresh}
\mathrm{soft}_{\lambda}\left( a\right)&=
\left\{
\begin{array}{ll}
a- \lambda & \mbox{if $a \geq \lambda$,}\\
a+ \lambda & \mbox{if $a\leq - \lambda$,}\\
0 & \mbox{otherwise.}
\end{array}
\right. 
\end{align} 
Let us make the following remarks about the different steps of the ADMM recursion. 
First, problem \eqref{eq:step1_ADMM} is a differentiable but non-convex and $n$-dimensional  minimization  problem.  Second, \eqref{eq:step2_ADMM_prox} corresponds to the definition of the proximal operator of the $\ell_1$ norm and  is based on the simple soft-thresholding operator~\eqref{eq:softthresh}.  
We note that, the solution of \eqref{eq:step2_ADMM_prox} is typically sparse since  soft-thresholding  enforces the small coefficients to be equal to zero. 

Interestingly, we remark that the solution of the soft constrained problem   \eqref{eq:probDefSoft} can be computed using the same steps \eqref{eq:step1_ADMM}-\eqref{eq:step3_ADMM_prox}, but substituting the first update in  \eqref{eq:step3_ADMM_prox}  by $\mathbf{u}_{{\mathbf{d}}}^{(i+1)}=0$.

Note that since the steps \eqref{eq:step2_ADMM_prox}-\eqref{eq:step3_ADMM_prox} of the procedure only involves vector additions, the particularization of ADMM to our problem leads to an algorithm exhibiting a complexity $\mathcal{O}(Km)$ per iteration scaling linearly in the problem dimensions, as long as the first step~\eqref{eq:step1_ADMM}  scales linearly. We will see in the Section~\ref{sec:nonLinear} that the first step is solved in fact in a log linear time.
However,  it is important to notice that  step~\eqref{eq:step1_ADMM} is the  computational bottleneck  of the ADMM procedure, because, on the contrary  to the  other two ADMM steps,  the high-dimensional problem  can not be divided into lower-dimensional sub-optimization problems of complexity independent of $K$.   Indeed,  the $K$ subsets of variables  $\{{\boldsymbol{d}^k},\boldsymbol{c}^k\}$ are all inter-depending  through the action of vertical winds  $\boldsymbol{\omega}$. Thus,  this ADMM procedure  does not offer a natural  structure  for parallelization, which  constitute a crucial issue for this high-dimensional optimization problem.

 \subsection{Dealing with High-Dimensionality   by Splitting}\label{sec:splitting}
  As mentioned previously, the ADMM recursion \eqref{eq:step1_ADMM}-\eqref{eq:step3_ADMM_prox} gathers $K$  sets of high-dimensional variables, interacting  on each others  through the $(K-1)$ vertical wind  fields.  Strategies based on optimal control can be efficient to deal  with this dependence of variables across vertical levels~\cite{Bertsekas99}. However, the ADMM algorithm enjoys also a very high popularity as an efficient tool for the fast solution of large-scale optimization problems  due to its ability at taking advantage of the structure of the variable dependance through operator splitting~\cite{glowinski2017splitting}.  Assume  without loss of generality that $K$ is even.  As proposed  hereafter, a variable splitting allows to solve \eqref{eq:probDef} by solving independently  $K/2$ sub-problems, and thus enabling parallel computation of complexity independent of~$K$. 

Problem \eqref{eq:ADMMProb}
 can equivalently written as
 \begin{equation}\label{eq:ADMMProbSplit}
\left\{\begin{aligned}
%&(\mathbf{e}_{\boldsymbol{  d}} , \boldsymbol{\omega},\boldsymbol{\tilde \omega},\boldsymbol{h}, \boldsymbol{\tilde h},\boldsymbol{c},\boldsymbol{\tilde c},\boldsymbol{\epsilon}) 
%=\\
& \argmin_{\mathbf{d},\boldsymbol{\omega},\boldsymbol{\tilde \omega},\boldsymbol{c},\boldsymbol{\tilde c}} \mathcal{J}(\mathbf{d}, \boldsymbol{\omega},\boldsymbol{c})+ \boldsymbol{\alpha_d}\mathcal{R}_d({\boldsymbol{d}} )+ \boldsymbol{\alpha}_{\xx}\mathcal{R}_\xx(\boldsymbol{\tilde c}),\\
&\textrm{s.t.}\quad 
\begin{pmatrix} \mathbf{D}&- \mathbf{L}&\mathbf{0}\\  \mathbf{0} & \mathbf{I} & \mathbf{0} \\  \mathbf{0} & \mathbf{0} &\mathbf{I} \end{pmatrix}\begin{pmatrix}{\boldsymbol{ d}}\\ \boldsymbol{\omega}    \\\boldsymbol{c}\end{pmatrix} 
+ \begin{pmatrix}  \mathbf{0}& \mathbf{0}\\ -\mathbf{I}&  \mathbf{0}\\  \mathbf{0}&  -\mathbf{I}\\ \end{pmatrix}\begin{pmatrix} \boldsymbol{\tilde \omega}    \\ \boldsymbol{\tilde c} \end{pmatrix} 
=\begin{pmatrix} \mathbf{0}\\ \mathbf{0}\\ \mathbf{0}\end{pmatrix} ,
  \end{aligned}\right.
\end{equation}
where we have added a new variable to the problem \eqref{eq:ADMMProb}: $\boldsymbol{\tilde \omega}$ made of the concatenation of respectively vectors $\boldsymbol{\tilde \omega}^1,...,\boldsymbol{\tilde \omega}^{K-1}$, which is counterbalanced by the inclusion of the  new constraint $\boldsymbol{\tilde \omega}=\boldsymbol{ \omega}$.

 As detailed in Appendix \ref{app:Bbis}, by splitting the set of variables  into  the two sub-sets according to their height level indexes whether in $\mathcal{K}'=\{0,2,4,\ldots,K-2\}$ or in $\mathcal{K}''=\{1,3,5,\ldots,K-1\}$,  the constrained optimization problem~\eqref{eq:ADMMProbSplit} fits the generic ADMM  framework exposed in appendix \ref{app:A}.  
 
 Let 
 \begin{align*}
 \mathcal{F}{({{\mathbf{d}}^k},{\boldsymbol{\omega}^k}, {\boldsymbol{\omega}^{k+1}} ,{\boldsymbol{c}^k})} &=
 \frac{\alpha_d^k}{2} \|  {\boldsymbol{\Delta}_m}\mathbf{d}^k\| ^2_2 + \frac{1}{2} \|\boldsymbol{\delta}({\boldsymbol{ \omega}^k}, {\boldsymbol{ \omega}^{k+1}})\|_2^2 \\
  &+ \frac{\rho}{2}\| \delta p^k\textrm{div}\,\mathbf{d}^k-{\boldsymbol{\omega}^{k}}+{\boldsymbol{\omega}^{k+1}}      
  + \mathbf{u}_{\mathbf{{\mathbf{d}^k}}}^{(i)} \|^2_2+\frac{\rho}{2} \|  \mathbf{c}^k -   {\mathbf{\tilde c}^k} \,^{(i)}+  \mathbf{u}_{\mathbf{c}^k}\,^{(i)} \|^2_2  ,
\end{align*}
and let 
\begin{align*}
\mathcal{G}{({{\mathbf{d}}^k},{\boldsymbol{\omega}^k}, {\boldsymbol{\omega}^{k+1}} ,{\boldsymbol{c}^k})}&=
\mathcal{F}{({{\mathbf{d}}^k},{\boldsymbol{\omega}^k}, {\boldsymbol{\omega}^{k+1}} ,{\boldsymbol{c}^k})}  + \underset{k'\in\{k,k+1\}}{\sum}  \frac{\rho}{2} \|  \boldsymbol{\omega}^{k'} -   { \boldsymbol{\tilde \omega}^{k'}} \,^{(i)}+  {\mathbf{u}_{\boldsymbol{ \omega}^{k'}}}^{(i)} \|^2_2,\\
  \mathcal{\tilde G}{({{\mathbf{d}}^k},{\boldsymbol{\tilde \omega}^k}, {\boldsymbol{\tilde \omega}^{k+1}} ,{\boldsymbol{c}^k})}&=
\mathcal{F}{({{\mathbf{d}}^k},{\boldsymbol{\tilde \omega}^k}, {\boldsymbol{\tilde \omega}^{k+1}} ,{\boldsymbol{c}^k})}  + \underset{k'\in\{k,k+1\}}{\sum}  \frac{\rho}{2} \|  {\boldsymbol{\omega}^{k'}}^{(i+1)} -   { \boldsymbol{\tilde \omega}^{k'}} +  \mathbf{u}_{\boldsymbol{ \omega}^{k'}}\,^{(i)} \|^2_2,
% &\mathcal{\tilde G}{({\mathbf{e}_{\mathbf{d}}^k},{\boldsymbol{\tilde \omega}^k}, {\boldsymbol{\tilde \omega}^{k+1}},{\boldsymbol{h}^k} ,{\boldsymbol{c}^k} ,{\boldsymbol{\epsilon}^k})} =
%\mathcal{F}{({\mathbf{e}_{\mathbf{d}}^k},{\boldsymbol{h}^k} ,{\boldsymbol{c}^k} ,{\boldsymbol{\epsilon}^k})} + \underset{\xx\in\{q,T,o\}}{\sum}\frac{\kappa_\xx}{2} \zeta \|\boldsymbol{\delta}_{\xx}^k({\boldsymbol{\tilde \omega}^k}, {\boldsymbol{\tilde \omega}^{k+1}})\|_2^2   
% \\
% &+\frac{\rho}{2}\| \delta p^k\textrm{div}\,\boldsymbol{\bar \Psi}_m\,{\mathbf{e}^k_\mathbf{d}}-{\boldsymbol{\tilde \omega}^{k}}+{\boldsymbol{\tilde \omega}^{k+1}}  + \mathbf{u}_{\mathbf{e^k_{\mathbf{d}}}}^{(i)} \|^2_2,
 \end{align*}
where we have  noted the data term $\frac{1}{2} \|\boldsymbol{\delta}(\boldsymbol{\theta},\yy)\|^2_2$ defined  in~\eqref{eq:likelyGauss} as $\frac{1}{2} \|\boldsymbol{\delta}({\boldsymbol{ \omega}^k}, {\boldsymbol{ \omega}^{k+1}})\|_2^2$, in order to simplify notations and to explicitly show  its dependance on the vertical winds $({\boldsymbol{ \omega}^k}, {\boldsymbol{ \omega}^{k+1}})$ or $({\boldsymbol{\tilde \omega}^k}, {\boldsymbol{\tilde \omega}^{k+1}})$ depending if $k$ belongs to $\mathcal{K}'$ or $\mathcal{K}''$. Using the proposed splitting, a local minimum of problem~\eqref{eq:ADMMProbSplit} can be obtained using ADMM  by iterating until convergence the 
  three following  steps 
 \begin{align}
 &\left\{
 \begin{array}{ll}
 &({{\mathbf{d}}^k}^{(i+1)},{\boldsymbol{\omega}^k}^{(i+1)}, {\boldsymbol{\omega}^{k+1}}^{(i+1)},{\boldsymbol{c}^k}^{(i+1)} ) =\underset{{{\mathbf{d}}^k},{\boldsymbol{\omega}^k}, {\boldsymbol{\omega}^{k+1}} ,{\boldsymbol{c}^k}}{\argmin}\,     \mathcal{G}{({{\mathbf{d}}^k},{\boldsymbol{\omega}^k}, {\boldsymbol{\omega}^{k+1}} ,{\boldsymbol{c}^k})}, \,k\in \mathcal{K}' \\
&({{\mathbf{d}}^k}^{(i+1)},{\boldsymbol{\tilde \omega}^k}^{(i+1)}, {\boldsymbol{\tilde \omega}^{k+1}}^{(i+1)},{\boldsymbol{c}^k}^{(i+1)} ) =\underset{{{\mathbf{d}}^k},{\boldsymbol{\tilde \omega}^k}, {\boldsymbol{\tilde \omega}^{k+1}} ,{\boldsymbol{c}^k}}{\argmin}\,     \mathcal{\tilde G}{({{\mathbf{d}}^k},{\boldsymbol{\tilde \omega}^k}, {\boldsymbol{\tilde \omega}^{k+1}} ,{\boldsymbol{c}^k})}, \,k\in \mathcal{K}'' 
%&\bullet\,\textrm{for
% }\, k\in \mathcal{K}'',\,({\mathbf{e}_{\mathbf{d}}^k}^{(i+1)},{\boldsymbol{\tilde \omega}^k}^{(i+1)}, {\boldsymbol{\tilde \omega}^{k+1}}^{(i+1)},{\boldsymbol{h}^k}^{(i+1)} ,{\boldsymbol{c}^k}^{(i+1)} ,{\boldsymbol{\epsilon}^k}^{(i+1)}) =\\
%  &\hspace{2cm}\underset{{\mathbf{e}_{\mathbf{d}}^k},{\boldsymbol{\tilde \omega}^k}, {\boldsymbol{\tilde \omega}^{k+1}},{\boldsymbol{h}^k} ,{\boldsymbol{c}^k} ,{\boldsymbol{\epsilon}^k}}{\argmin}\,    \mathcal{\tilde G}{({\mathbf{e}_{\mathbf{d}}^k},{\boldsymbol{\tilde \omega}^k}, {\boldsymbol{\tilde \omega}^{k+1}},{\boldsymbol{h}^k} ,{\boldsymbol{c}^k} ,{\boldsymbol{\epsilon}^k})} \\
 \end{array}
 \right.,\label{eq:step1_ADMMSplitMore}\\
&{\mathbf{\tilde c}^{k,\ell}\,}^{(i+1)}(s)=\mathrm{soft}_{\frac{{\alpha}^k_{\xx}}{\rho}} \left(\mathbf{c}^{k,\ell}\,^{(i+1)}(s) + \mathbf{u}^{k,\ell}_{\boldsymbol{c}}\,^{(i)}(s)\right),
  k\in \mathcal{K}, \ell\in\{1,2,3\},  s:\chi(s)\in \Omega_m ,\label{eq:step2_ADMMSplitMore}\\
% &\left\{
% \begin{array}{ll}
% \mathbf{\tilde h}^k\,^{(i+1)}(s)&=\mathrm{soft}_{\frac{\alpha_h}{\rho}} \left(\mathbf{h}^k\,^{(i+1)}(s) + \mathbf{u}^k_{\boldsymbol{h}}\,^{(i)}(s)\right)\label{eq:step2_ADMMSplitMore}\\
% \mathbf{\tilde c}^k_\mathbf{x}\,^{(i+1)}(s)&=\mathrm{soft}_{\frac{\boldsymbol{\alpha}_{\xx}}{\rho}} \left(\mathbf{c}_\mathbf{x}^k\,^{(i+1)}(s) + \mathbf{u}^k_{\boldsymbol{c}_\mathbf{x}}\,^{(i)}(s)\right)
%\end{array}\quad    k\in \mathcal{K},\, 1\le s  \le  n ,
% \right.\\
%&\left\{
% \begin{array}{ll}
%\boldsymbol{\tilde h'}^{(i+1)} = \argmin_{\boldsymbol{\tilde h'}}  \|  \boldsymbol{\tilde h^{'}}  \|_1 + \frac{\rho}{2 \alpha_h} \|\boldsymbol{ h^{'}}^{(i+1)} - \boldsymbol{\tilde h}  + \mathbf{u}_{\boldsymbol{h}^{'}}^{(i)} \|^2_2, \\
%\boldsymbol{\tilde h^{''}}^{(i+1)} = \argmin_{\boldsymbol{\tilde h^{''}}}  \|  \boldsymbol{\tilde h^{''}}  \|_1 + \frac{\rho}{2 \alpha_h} \|\boldsymbol{ h^{''}}^{(i+1)} - \boldsymbol{\tilde h^{''}}  + \mathbf{u}_{\boldsymbol{h}^{''}}^{(i)} \|^2_2, \\
%\boldsymbol{\tilde c'}^{(i+1)} = \argmin_{\boldsymbol{\tilde c'}}  \|  \boldsymbol{\tilde c'}  \|_1 + \frac{\rho}{2 \boldsymbol{\alpha}_{\xx}} \|\boldsymbol{ c'}^{(i+1)} - \boldsymbol{\tilde c'}  + \mathbf{u}_{\boldsymbol{c'}}^{(i)} \|^2_2 ,\\ 
%\boldsymbol{\tilde c^{''}}^{(i+1)} = \argmin_{\boldsymbol{\tilde c^{''}}}  \|  \boldsymbol{\tilde c^{''}}  \|_1 + \frac{\rho}{2 \boldsymbol{\alpha}_{\xx}} \|\boldsymbol{ c^{''}}^{(i+1)} - \boldsymbol{\tilde c^{''}}  + \mathbf{u}_{\boldsymbol{c}^{''}}^{(i)} \|^2_2 , \\
% \end{array}
% \right.\\
 &\left\{
 \begin{array}{ll}
&\mathbf{u}_{{{\mathbf{d}^k}}}^{(i+1)}= \mathbf{u}_{{{\mathbf{d}^k}}}^{(i)} + \delta p^k\textrm{div}{\mathbf{d}^k}^{(i+1)}-{\boldsymbol{\omega}^{k}}^{(i+1)}+{\boldsymbol{\omega}^{k+1}}^{(i+1)},  \quad  k\in \mathcal{K}', \\
&\mathbf{u}_{{{\mathbf{d}^k}}}^{(i+1)}= \mathbf{u}_{{{\mathbf{d}^k}}}^{(i)} + \delta p^k\textrm{div}{\mathbf{d}^k}^{(i+1)}-{\boldsymbol{\tilde \omega}^{k}}^{(i+1)}+{\boldsymbol{\tilde \omega}^{k+1}}^{(i+1)},  \quad  k\in \mathcal{K}'',\\
&\mathbf{u}_{\boldsymbol{c}}^{(i+1)}\,\,\,\,\,\, = \mathbf{u}_{\boldsymbol{c}}^{(i)} + {\boldsymbol{c}}^{(i+1)} -{\boldsymbol{\tilde c}}^{(i+1)},\label{eq:step3_ADMMSplitMore}\\
 &\mathbf{u}_{\boldsymbol{\omega}^k}^{(i+1)} = \mathbf{u}_{\boldsymbol{\omega}^k}^{(i)} + {\boldsymbol{\omega}^k}^{(i+1)} -{\boldsymbol{\tilde \omega}^k}^{(i+1)}, \quad  k\in \mathcal{K}.
%&\mathbf{u}_{\boldsymbol{c}^k_\xx}^{(i+1)} =\mathbf{u}_{\boldsymbol{c}^k_\xx}^{(i)} + {\boldsymbol{c}^k_\xx}^{(i+1)} -{\boldsymbol{\tilde c}^k_\xx}^{(i+1)},\quad  k\in \mathcal{K}'\cup\mathcal{K}'', \quad \xx \in \{q,T,o\},\\
\end{array}
 \right.
\end{align}

The $K/2$ minimization problems can be performed independently  within the two consecutive sub-steps appearing  in system  \eqref{eq:step1_ADMMSplitMore}.  In other words, this independence allows to perform sequentially  the two sub-steps, each one of them  constituted of $K/2$ parallel minimization procedures. The other two steps, namely \eqref{eq:step2_ADMMSplitMore} and \eqref{eq:step3_ADMMSplitMore} can be performed independently for the different indexes $k\in \mathcal{K}=\mathcal{K}'\cup\mathcal{K}''$.
% Interaction between the   $K$ recursions occur in the ADMM step  \eqref{eq:step3_ADMMSplit}. In particular interaction is taken into account in  the first equation, which  performs an ascent step of the dual function whose gradient is  the constraint on the splitting of the vertical wind variables:  $\boldsymbol{\omega}-\boldsymbol{\tilde \omega}$.
 Therefore, the proposed scheme enables to solve \eqref{eq:probDef} with an overall computation time $\mathcal{O}(m)$ independent of  $K$, as long as we are able to solve $K/2$ minimization problem in parallel.

As for the  solver proposed in Section~\ref{sec:DealingNonDiff}, we notice that the solution of the soft constrained problem   \eqref{eq:probDefSoft} can be computed using the steps \eqref{eq:step1_ADMMSplitMore}-\eqref{eq:step3_ADMMSplitMore}, but substituting the two first updates in  \eqref{eq:step3_ADMMSplitMore}  by $\mathbf{u}_{{\mathbf{d}^k}}^{(i+1)}=0$.

  \subsection{Dealing with Non-Convexity}
    \subsubsection{ADMM convergence}
 The convergence of the ADMM algorithm to a local minimum is guaranteed for a non-convex objective function under mild conditions as detailed in~\cite{wang2019global}. In the context of our optimization problem \eqref{eq:probDef}, a sufficient assumption is that the objective function is coercive. Due to the regularizer \eqref{eq:regC} and the linear constraint in \eqref{eq:probDef}, it is straightforward to show the coercivity with respect to the variables $\boldsymbol{\omega}$ and $\xx_{t_1}$, for fixed  $\mathbf{d}$. However, the coercivity with respect to $\mathbf{d}$ is not guaranteed with a regularizer of the form of \eqref{eq:DefR2}. To verify the coercivity assumption, one may add to the cost function a Tikonov regularizer  term of the form $\delta \|\mathbf{d}\|_2^2$ with $\delta$ being a small positive constant, set above the machine precision.
 \subsubsection{Efficient  Optimization} \label{sec:nonLinear}
 The two parallel minimizations in \eqref{eq:step1_ADMMSplitMore} are non-convex but differentiable problems. They may be solved efficiently using gradient-descent methods, as for instance the  limited-memory quasi-Newton descent method, known as  Broyden-Fletcher-Goldfarb-Shanno (L-BFGS) procedure, with a line-search routine  based on the strong Wolf conditions \cite{Nocedal99}.   Such descent methods involve a complexity per iteration scaling linearly  in $m$,  to which must be added the complexity to compute the gradient of the objective in~\eqref{eq:step1_ADMMSplitMore}. Let us mention that,  interestingly,  the minimizations in \eqref{eq:step1_ADMMSplitMore} need not (under very mild conditions) to be  performed exactly to guaranteethe convergence of ADMM, see \eg,  \cite[Theorem 8]{Eckstein1992DouglasRachford}.

 Nevertheless, it is well known that large displacements are difficult to estimate due to the cost function non-linearities. To face this problem, as proposed in~\cite{derian2013wavelets} and latter studied in~\cite{Kadri13,Heas14,schmidt2019high,nicolas2023assessment}, the optimization procedure avoids the heuristic and standard multiresolution optic flow initialization, and relies instead on the estimation of wavelet  expansions of the displacement variable $\mathbf{d}$  (Coiflets with 10 vanishing moments). The strategy consists in estimating wavelet series with an increasing number of terms over the iterations of the algorithm, the added terms being related to increasingly fine spatial scales. The wavelet decomposition is in turn exploited to make the computation of the  Laplacian operator efficient   in the regularizer \eqref{eq:DefR2}~\cite{Kadri13}.  An analogous wavelet expansion is used to extend the image variables $\xx_{t_1}$ accordingly. The fast evaluation procedures  computing the  gradient of the cost function is  similar to the one proposed in~\cite{Heas14}. The fast evaluation  relies on the fast wavelet transform  and on the fast Fourier transform computed in log linear time. In summary, the complexity of the overall algorithm is $\mathcal{O}(m \log m)$.

\begin{figure}[!t]
\vspace{-0.cm}\begin{center}
\begin{tabular}{cc||cc}
\multicolumn{2}{c}{\footnotesize  {  ECMWF  model  }}&\multicolumn{2}{c}{\footnotesize  {  IASI  observations (synchronized)}}\\
\hline \\
\hspace{-0.75cm}\rotatebox{90}{\parbox{2mm}{{{\footnotesize{$\quad\quad ({\mathbf{d}^k}^{\star},{{\boldsymbol{\omega}^k}^\star})$}}}}}\hspace{0.25cm}\includegraphics[width=0.23\textwidth]{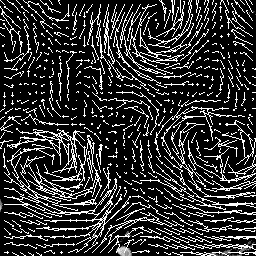}&\hspace{-0.25cm}\includegraphics[width=0.23\textwidth]{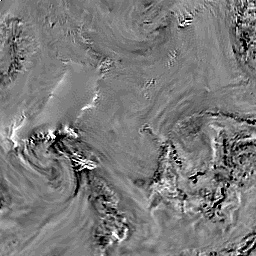}&\\
\hspace{-0.5cm}\rotatebox{90}{\parbox{2mm}{{{}}}}\hspace{0.25cm}\includegraphics[width=0.23\textwidth]{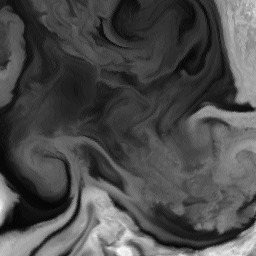}&\hspace{-0.25cm}\includegraphics[width=0.23\textwidth]{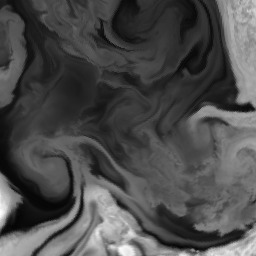}&\hspace{-0.cm}\includegraphics[width=0.23\textwidth]{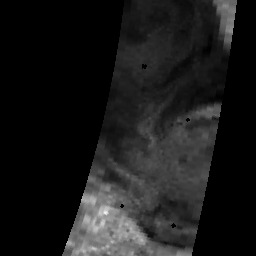}&\hspace{-0.25cm}\includegraphics[width=0.23\textwidth]{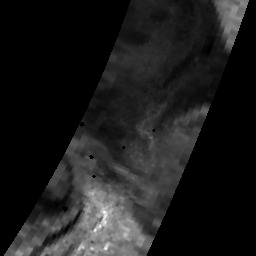}\\
\hspace{-0.75cm}\rotatebox{90}{\parbox{2mm}{{{\footnotesize{$\quad\quad  (\mathbf{y}^{k}_{t_0},\, \mathbf{y}^{k}_{t_1})$}}}}}\hspace{0.25cm}\includegraphics[width=0.23\textwidth]{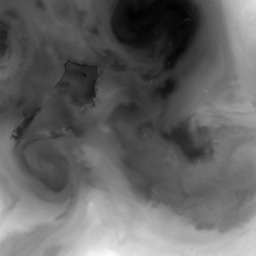}&\hspace{-0.25cm}\includegraphics[width=0.23\textwidth]{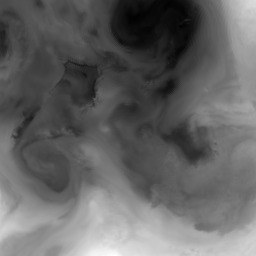}&\hspace{-0.cm}\includegraphics[width=0.23\textwidth]{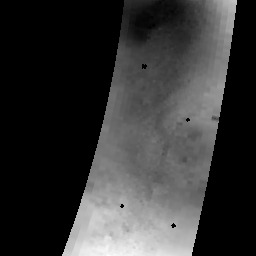}&\hspace{-0.25cm}\includegraphics[width=0.23\textwidth]{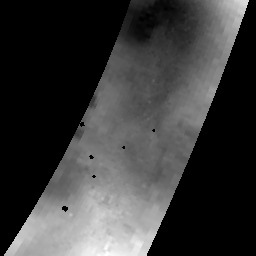}\\
\hspace{-0.5cm}\rotatebox{90}{\parbox{2mm}{}}\hspace{0.25cm}\includegraphics[width=0.23\textwidth]{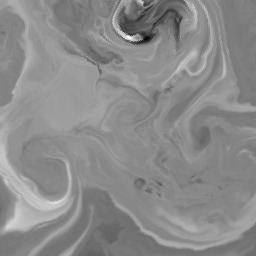}&\hspace{-0.25cm}\includegraphics[width=0.23\textwidth]{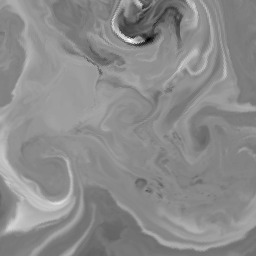}&\hspace{-0.cm}\includegraphics[width=0.23\textwidth]{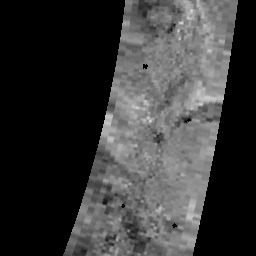}&\hspace{-0.25cm}\includegraphics[width=0.23\textwidth]{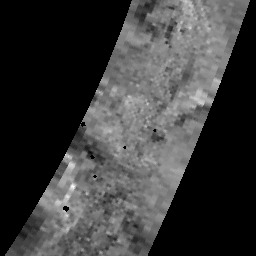}\\
\footnotesize{$t_0$}&\footnotesize{$t_1$}&\footnotesize{$t_0$}&\footnotesize{$t_1$}\vspace{-0.2cm}
\end{tabular}
	\caption{{\footnotesize  {  Horizontal AMVs at altitude 800-850 hPa with the vertical wind at the lower boundary (800 hPa) from the ECMWF  simulation (first row). Pairs of image observations (3 lats rows): humidity (above), temperature (middle), ozone concentration (below) for ECMWF or IASI datasets (black pixels correspond to missing data).  \label{fig:1}}}}\vspace{-0.cm}
		\end{center}\vspace{-0.5cm}
\end{figure}

\section{Numerical Evaluation}\label{sec:NumEval} 
\subsection{Data Benchmark}

We detail hereafter two different  datasets issued from   real-world meteorological observations.

  \begin{itemize}
  \item {\bf ECMWF Data.} 
  Observations
are provided, together with the corresponding ground truth profiles of 3D motion
fields, by the operational numerical model of the European Centre for Medium-Range Weather Forecasts (ECMWF) \cite{temperton2001two}. There are neither noise nor missing image observations.
  
   \item {\bf IASI Data.} 
Incomplete observations are provided by the Infrared Atmospheric Sounding Interferometer (IASI) of Metop-A and Metop-B satellites~\cite{borde2019winds}, while a proxy to the
ground truth is assumed to be the synchronized ECMWF numerical model. The exploitation of this dataset for the estimation of vertical profiles of dense 3D AMV fields is extremely ambitious because there are many missing observations in the images and in addition the remaining observations are highly corrupted by noise.
  \end{itemize}

In these two datasets, observations are composed of two consecutive vertical profiles of 3 physical quantities, namely pressure-averaged atmospheric humidity, temperature and ozone concentration. Observations are gathered in two stacks of  layers of tri-variate (possibly incomplete) images.  Altitudes   ranging from the isobaric levels   1000 hPa to 400 hPa are discretize in $K=8$ layers. {Specifically, the 8 pressure-averaged  humidity, temperature and ozone concentration images are integrals \eqref{eq:rhoDef} computed for height intervals $(p^{k+1},p^k) \in \{(1000,950),(950,900),(900,850),(850,800),(800,700),$ $(700,600),(600,500),(500,400)\}$, where the integrand is a piecewise constant approximation obtained using the 137 image bands provided by ECMWF or IASI}. Each image is of dimension $m = 256 \times 256 $.  

The IASI data gathers incomplete observation maps, on the contrary to the ECMWF data which provides dense observations. In consequence, the IASI dataset will require the joint estimation of the image variable $\xx_{t_1}$ and the 3D AMVs $(\mathbf{d},\boldsymbol{\omega})$, while only the latter couple of variables will need to be estimated in the case of the ECMWF dataset.  

As previously mentioned, in the context of the ECMWF data, we have the true parameter $\boldsymbol{ \theta}^\star$, which generated the observation $\mathbf{y}$. In the context of the IASI data, the synchronized ECMWF data will be considered as a proxy for the true parameter $\boldsymbol{ \theta}^\star$. We specify that this ground truth will be used exclusively for the analysis of the performance of the algorithms, and will not be used as an input ingredient of the proposed method.

\subsection{Algorithm Benchmark \&  Evaluation Criteria}\label{sec:algoBench}

We propose to compare the proposed solvers for problem  \eqref{eq:probDef} and \eqref{eq:probDefSoft},  with three state-of-the-art algorithms.
\begin{itemize}
\item {\bf 2D model.} This algorithm estimates horizontal 2D AMVs, solving problem \eqref{eq:probDefSoft} with the vertical winds  held at zero  ${\boldsymbol{\omega}}=0$ and with $\rho=0$, \ie, without the constraint \eqref{eq:linearConst}. This estimator constitutes our reference method: 2D optic flow estimation~\cite{derian2013wavelets}, performed independently on each atmospheric layer, coupled to the reconstruction of missing image observations, in the spirit of the work~\cite{heas2016efficient}. 
\item {\bf 2D  incompressible model (soft constraint).} This algorithm estimates horizontal 2D AMVs, solving problem \eqref{eq:probDefSoft} with the vertical wind vector  held at zero  ${\boldsymbol{\omega}}=0$.   In this zero-vertical wind setting, \eqref{eq:linearConst} boils down to a soft zero-divergence constraint. This estimator constitutes an alternate version of the reference 2D model, where we have added a divergence-free constraint, similar to~\cite{Kadri13,Heas14}. 
\item {\bf 3D model.} This algorithm estimates 3D AMVs, solving problem \eqref{eq:probDefSoft}  with $\rho=0$, \ie, without the hydrostatic constraint \eqref{eq:linearConst}. This estimator implements an enhanced version of the  3D model proposed in~\cite{heas2008three}, in which motion estimation is coupled to  the reconstruction of missing observations\footnote{In addition to the difference in the 3D model, our solver also differs from the algorithm of~\cite{heas2008three} by the use of an efficient high-dimensional optimization procedure based on ADMM splitting (see Section~\ref{sec:splitting}) and efficient minimization using the well-known L-BFGS procedure and employing a wavelet-based multiresolution framework (see Section~\ref{sec:nonLinear}).}.
\item {\bf 3D  hydrostatic model (soft constraint).} This algorithm implements the proposed solver for problem \eqref{eq:probDefSoft}.
\item {\bf 3D  hydrostatic  model (hard constraint).} This algorithm  implements the proposed  solver for  problem \eqref{eq:probDef}.

\end{itemize}   

The free parameters of the warping constraint \eqref{eq:splineRepr} and the hydrostatic constraint \eqref{ICE2}, namely the $\gamma^k$'s and the $\delta p^k$'s, are learned in a minimum mean square sense from the ECMWF data. The regularization parameters $\boldsymbol{\alpha}_{\mathbf{d}}$, $\boldsymbol{\alpha}_{\xx}$ and  $\rho$ in the cost \eqref{eq:probDef} or \eqref{eq:probDefSoft} are set on the basis of expert knowledge. We mention that the algorithms all started by initializing the unknown AMV and image fields to zero. \\

A common criterion  to compare  the accuracy of the horizontal AMV components estimated at  layer $k$,  which we will denote by vector $   \mathbf{\hat d^k}$,  is the normalized average horizontal endpoint error (EPE)~\cite{Butler:ECCV:2012} 
\begin{align}\label{eq:HEPE}
 \textrm{Horizontal\, EPE}(k)=\frac{\sum_{j:\varkappa(j)\in\Omega_{obs}}\| \mathbf{d^k}^\star(j)-   \mathbf{\hat d^k}(j) \|_2}{\sum_{j:\varkappa(j)\in\Omega_{obs}}\|\mathbf{d^k}^\star(j)\|_2},
 \end{align}
 where $\Omega_{obs}$ stands here and in the following for the pixel grid with consecutive observations, \ie,  $\Omega_{obs}=\Omega^{t_0,k}_{obs} \cap  \Omega^{t_1,k}_{obs} $. 
 To fully assess the accuracy of 3D AMV estimation, we complement this criterion with the normalized root mean square error (RMSE) on vertical winds
\begin{align}\label{eq:VEPE} \textrm{Vertical\,RMSE}(k)=\left(\frac{\sum_{j:\varkappa(j)\in\Omega_{obs}}| \boldsymbol{\omega^k}^\star(j)-  \boldsymbol{\hat \omega^k}(j) |^2}{\sum_{j:\varkappa(j)\in\Omega_{obs}}|\boldsymbol{\omega^k}^\star(j)|^2}\right)^{1/2},
\end{align}
 where $\boldsymbol{\hat \omega^k}$ denotes the vector of vertical wind  estimates at layer $k$.

\begin{figure}[!t]
\begin{center}
\begin{tabular}{cc} 
\hline 
\multicolumn{2}{c}{\footnotesize{ECMWF}} \\
\hline  \vspace{-2.5cm}  \\
 \vspace{-2cm} \includegraphics[width=0.5\textwidth]{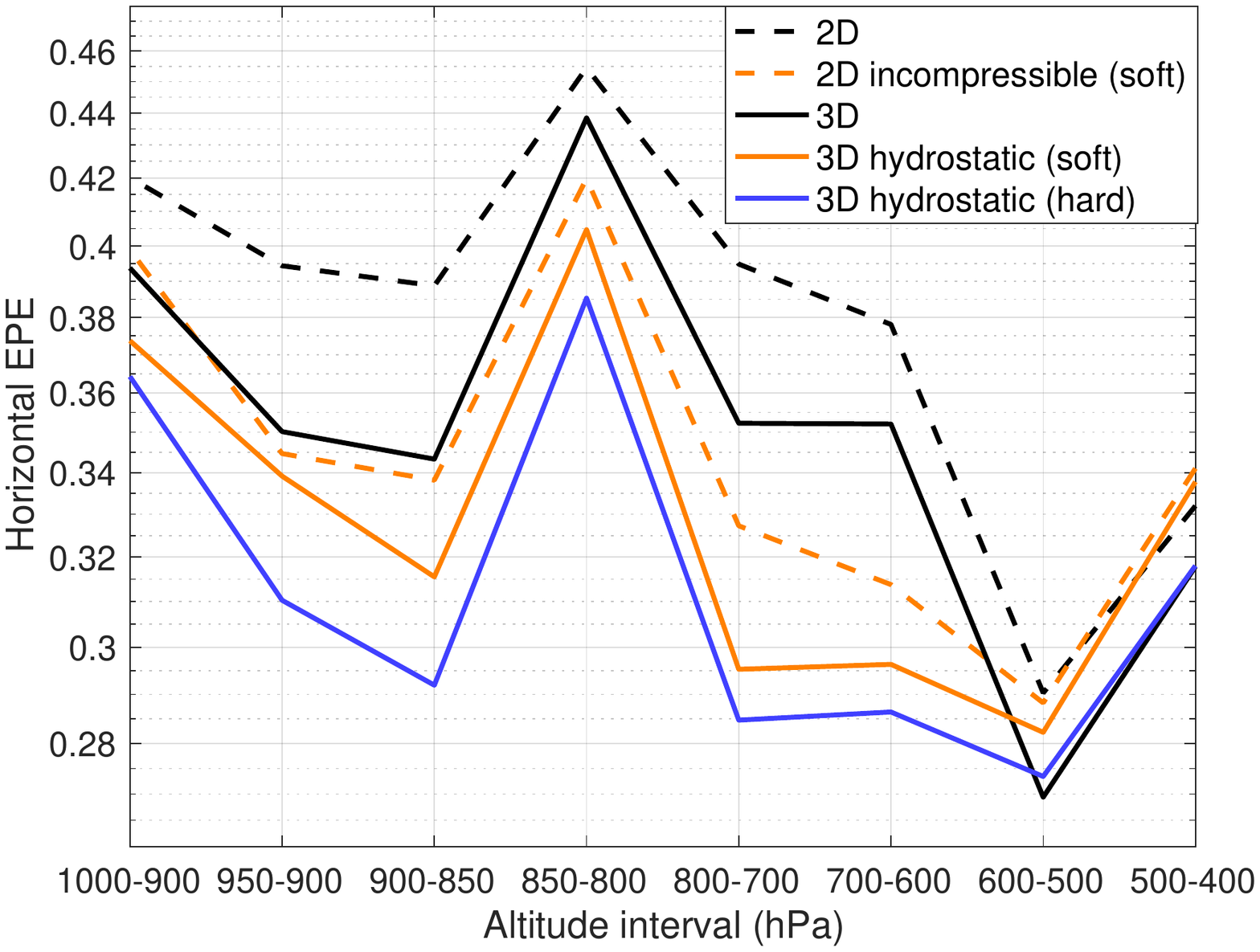}&\hspace{-0.25cm}\includegraphics[width=0.5\textwidth]{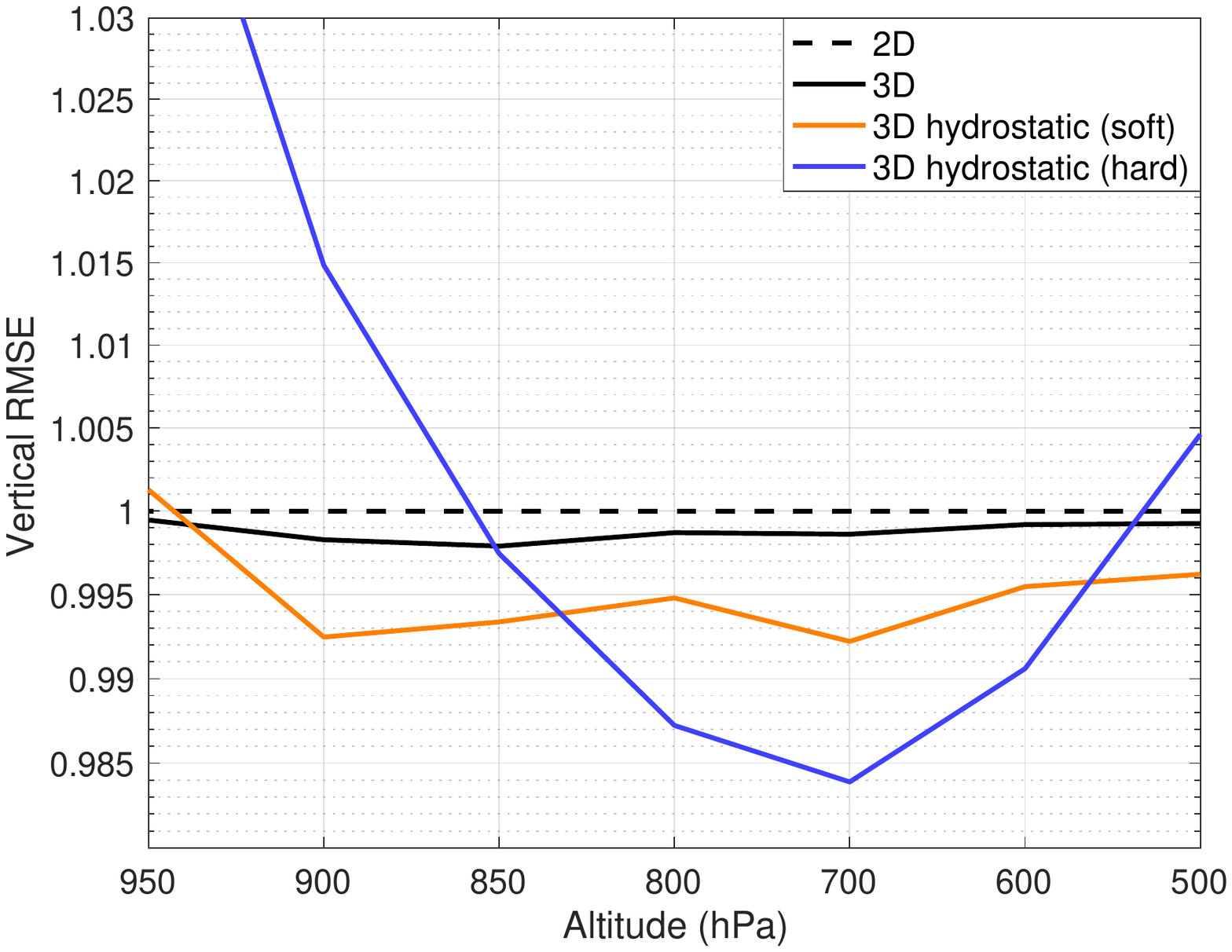} \vspace{-0cm}\\
 \hline 
\multicolumn{2}{c}{\footnotesize{IASI}} \\
\hline  \vspace{-2.5cm}  \\
\includegraphics[width=0.5\textwidth]{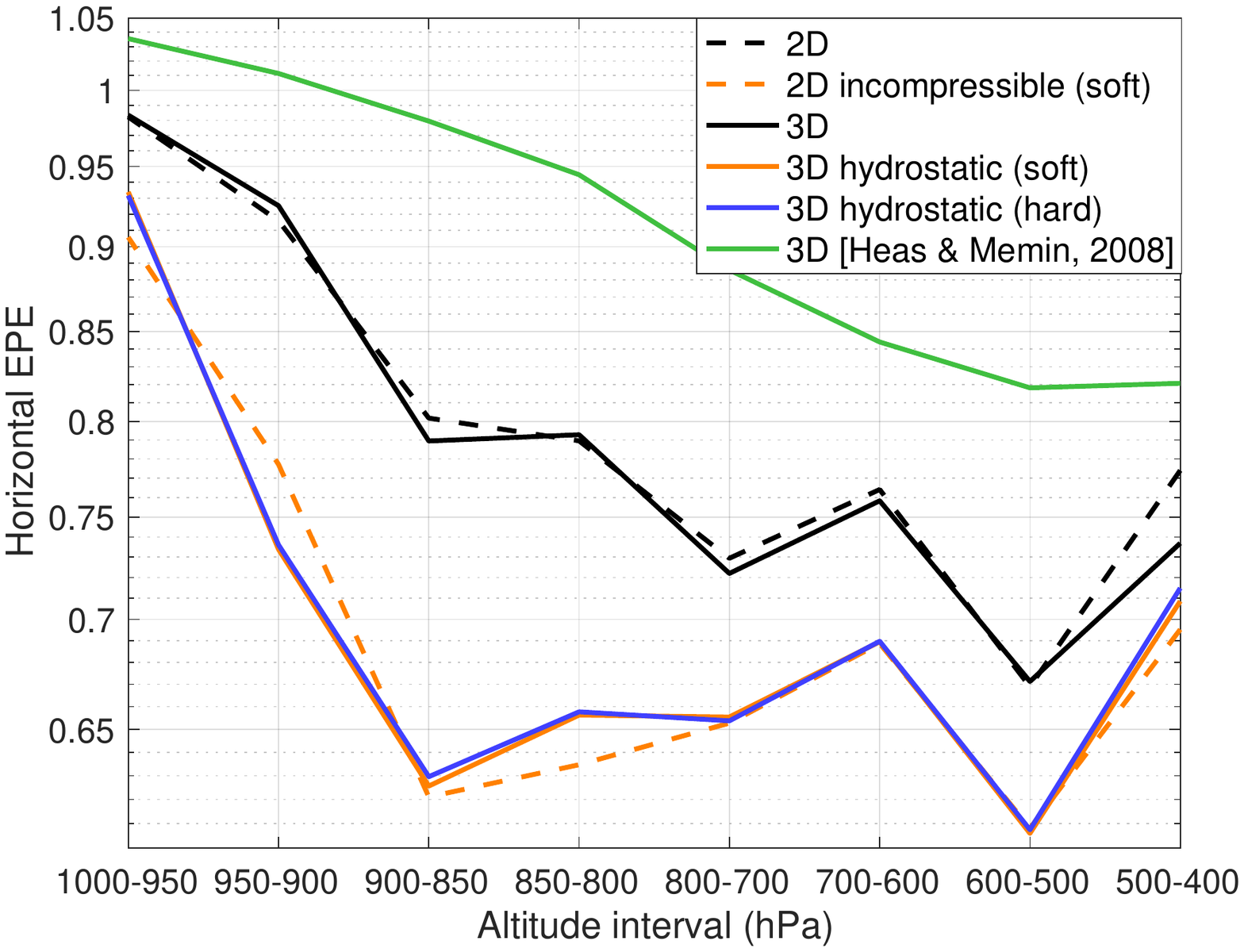}&\hspace{-0.25cm}\includegraphics[width=0.5\textwidth]{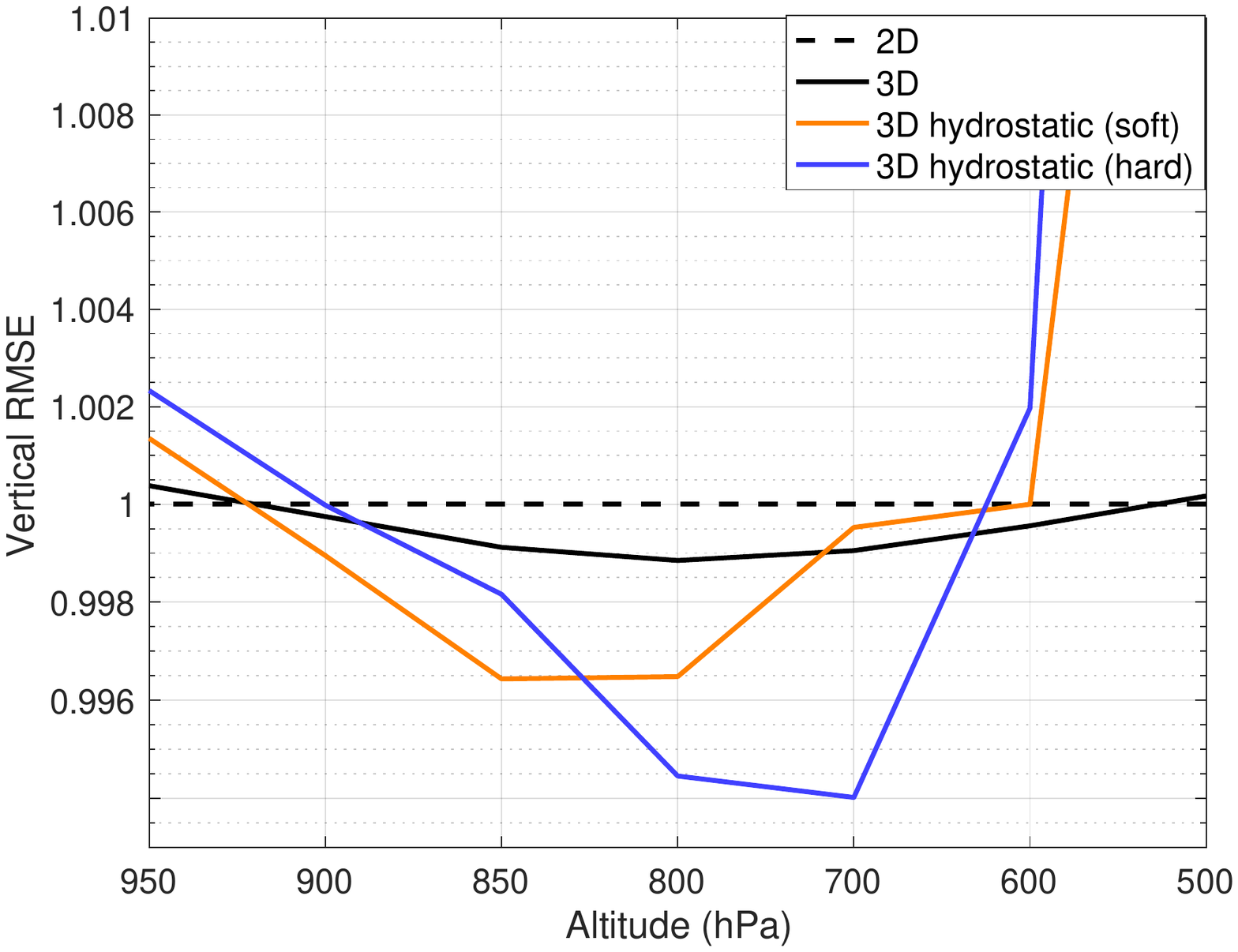}  \vspace{-2.5cm} 
\end{tabular}
	\caption{{\footnotesize  { Profiles of  error criteria \eqref{eq:HEPE} and \eqref{eq:VEPE} computed for the proposed and state-of-the-art algorithms  using the ECMWF or the IASI observations.\label{fig:2}}}}
		\end{center}\vspace{-0.2cm}
\end{figure}

\subsection{Results}

Figure~\ref{fig:2} plots the profiles of the error criteria \eqref{eq:HEPE} and \eqref{eq:VEPE} for the two datasets for the different algorithms of the benchmark   in Section~\ref{sec:algoBench}. 

\begin{figure}[!t]
\vspace{-0.cm}\begin{center}
\begin{tabular}{cc||cc}
\multicolumn{2}{c}{\footnotesize  {layer $k$ at 900-850 hPa  }}&\multicolumn{2}{c}{\footnotesize  {layer $k$ at 700-600 hPa}}\\
\hline \\
\hspace{-0.75cm}\rotatebox{90}{\parbox{2mm}{{{$\quad\quad\quad$\footnotesize{truth}}}}}\hspace{0.25cm}\includegraphics[width=0.23\textwidth]{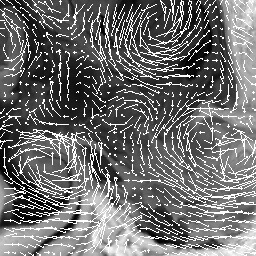}&\hspace{-0.25cm}\includegraphics[width=0.23\textwidth]{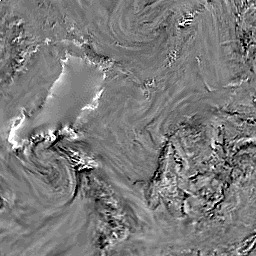}&\hspace{-0.cm}\includegraphics[width=0.23\textwidth]{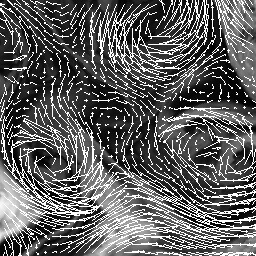}&\hspace{-0.25cm}\includegraphics[width=0.23\textwidth]{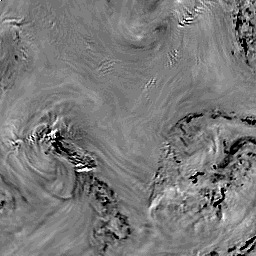}\\
\hspace{-0.75cm}\rotatebox{90}{\parbox{2mm}{{{$\quad\quad\quad$\footnotesize{2D}}}}}\hspace{0.25cm}\includegraphics[width=0.23\textwidth]{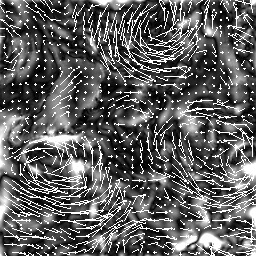}&&\hspace{-0.cm}\includegraphics[width=0.23\textwidth]{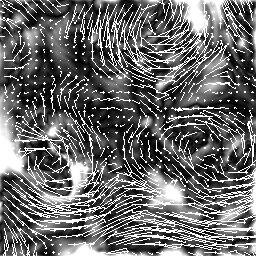}&\hspace{-0.25cm}\\
\hspace{-0.75cm}\rotatebox{90}{\parbox{2mm}{{{$\quad\quad\quad$\footnotesize{3D}}}}}\hspace{0.25cm}\includegraphics[width=0.23\textwidth]{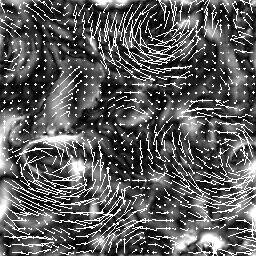}&\hspace{-0.25cm}\includegraphics[width=0.23\textwidth]{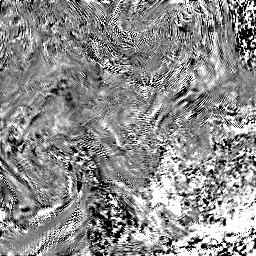}&\hspace{-0.cm}\includegraphics[width=0.23\textwidth]{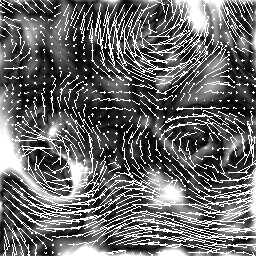}&\hspace{-0.25cm}\includegraphics[width=0.23\textwidth]{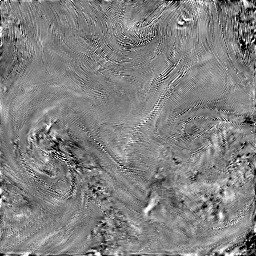}\\
\hspace{-0.75cm}\rotatebox{90}{\parbox{2mm}{{{$\quad\,\,$\footnotesize{3D$\,$hydro.$\,$(hard)}}}}}\hspace{0.25cm}\includegraphics[width=0.23\textwidth]{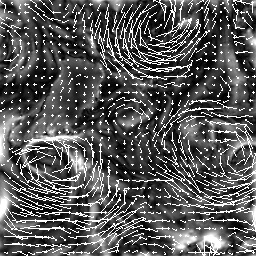}&\hspace{-0.25cm}\includegraphics[width=0.23\textwidth]{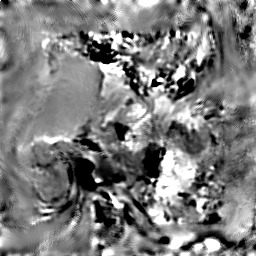}&\hspace{-0.cm}\includegraphics[width=0.23\textwidth]{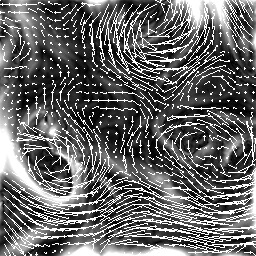}&\hspace{-0.25cm}\includegraphics[width=0.23\textwidth]{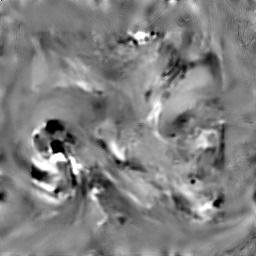}\\
{{\footnotesize{$ {\mathbf{d}^k}^{\star}$ plotted on $\mathbf{y}^{k}_{t_0}$ (above)  }}}&{{\footnotesize{$ {{\boldsymbol{\omega}^k}^\star}$ (above) or $  \boldsymbol{\hat \omega}^{k}$ }}}&{{\footnotesize{$ {\mathbf{d}^k}^{\star}$ (above) or $ {\mathbf{\hat d}^k}$ }}}&{{\footnotesize{$ {{\boldsymbol{\omega}^k}^\star}$ (above) or $  \boldsymbol{\hat \omega}^{k}$ }}}\\
{{\footnotesize{and $ {\mathbf{\hat d}^k}$ plotted on error }}}&&{{\footnotesize{plotted on error norm}}}&
\vspace{-0.cm}
\end{tabular}
	\caption{{\footnotesize  {Horizontal and vertical components of AMV fields estimated from the ECMWF data at two different altitudes.  Estimated vector fields $ {\mathbf{\hat d}^k}$ for the 2D, 3D or 3D hydrostatic models (three last rows) are superimposed on their related map of horizontal EPE whose $j$-th component is $\| \mathbf{d^k}^\star(j)-   \mathbf{\hat d^k}(j) \|_2$. The ground truth  vector fields $ {\mathbf{d}^k}^{\star}$ (row above) are superimposed on the related humidity field component of observations $\mathbf{y}^{k}_{t_0}$. \label{fig:3}}}}\vspace{-0.cm}
		\end{center}\vspace{-0.55cm}
\end{figure}

We focus on the upper plots, \ie, on the context of the noise-free and complete observations provided by the ECMWF numerical  weather  simulation.  We observe the relevance of both, the 3D versus the 2D models and the introduction of constraints (incompressibility for the 2D model or hydrostatic balance for the 3D ones). Specifically, accounting for interacting horizontal AMVs via vertical winds in a 3D model provides a systematic improvement of up to about 5\% in terms of normalized horizontal EPE.  We find that the 3D model also systematically decreases the vertical RMSE. However, this decrease is only of the order of a small fraction. Indeed, although the spatial structure of the estimated vertical wind fields is globally consistent with the ground truth (as can be seen in Figure~\ref{fig:3}), the vertical wind magnitude is dramatically underestimated, yielding a normalized RMSE slightly below unity. Next, the introduction of soft contraints on  incompressibility  in the 2D model or on hydrostatic balance in the 3D model lowers the error criteria a bit more, with a gain on the normalized horizontal EPE reaching up to 10\% for the constrained 3D model  compared to the reference 2D model. The hard constraint on hydrostatic balance finally yields the best  horizontal AMVs estimates in terms of EPE. We note that these constraints are relevant in terms of RMSE to estimates of the vertical wind away from Earth and the tropopause boundary layer, suggesting that the assumption of zero wind boundary conditions is somewhat spurious.

We continue with the lower plots, \ie,  the  noisy and incomplete observations measured by IASI. For this data context, we have at our disposal  state-of-the-art results obtained with the 3D model solver proposed in~\cite{heas2008three}. The latter constitutes the state-of-the-art algorithm to extract 3D AMVs from IASI data used  presently as a demonstration prototype   by the European operational satellite agency for monitoring of weather, climate and the environment from space (EUMETSAT). We immediately notice that the results of this state-of-the-art algorithm differ from those of our 3D model solver (see the~\ref{sec:algoBench} section and its footnote for more details on the differences of the model and solver).  Specifically, all the algorithms in our benchmark perform significantly better than the state-of-the-art method in terms of the two criteria (note that the corresponding vertical RMSE does not appear because it is well above unity). In this context of noisy and missing observations,  the horizontal EPE is obviously higher than for the ECMWF (about 40\% higher).   Commenting more specifically on the algorithms of our benchmark, we note that the performance of the 2D and 3D model are very similar in terms of horizontal EPE with a gain of about 15\% in comparison to state-of-the-art. A decrease reaching about 35\% is obtained  adding the incompressibility or hydrostatic (soft or hard) constraints. 
The significant improvement in accuracy  suggests that the noisy and incomplete observations supplemented by the a priori constraining divergence bursts are sufficient to characterize the dominant solenoidal structures of the flow. However, the small difference between the 2D and 3D model, i.e., little influence of the interaction by vertical winds of horizontal AMVs, suggests that, unlike the ECMWF dataset, the IASI observations are not rich enough to accurately characterize the weaker divergence structures (the magnitude of the solenoidal component is about twice in average that of the divergence) or the vertical component of the atmospheric flow. Nevertheless, we observe the same trend on the vertical RMSE profiles as for the ECMWF data, but in a smaller range. \\

After this quantitative evaluation, we move on to a qualitative analysis by comparing, for the different algorithms, the spatial structures of the estimated AMVs with those of the ground truth. 

In the case of the ECMWF data, Figure~\ref{fig:3} shows that the large vortex structures on the horizontal winds are accurately estimated by all algorithms while the errors at medium and small scales tend to disappear when moving from the 2D to the 3D model and then adding the hydrostatic balance constraint. We observe,  that the large scale structures of the vertical components of AMV are approximately  recovered by the 3D model, although they are embedded in strong noise. The hydrostatic balance constraint smoothes out this noise and reveal most of the large to medium structures of vertical winds. Although the results presented in Figure~\ref{fig:2} showed the positive impact of considering a 3D model on the accuracy of the horizontal wind estimates, the ability to capture the global spatial structures of the vertical wind was nevertheless not expected in view of the small gain in accuracy observed in the vertical RMSE profiles. 
 A potential explanation could be that the highly dimensional and strongly non-convex optimization at the heart of the proposed method remain stuck in local minima reflecting weak vertical wind states. This would preclude obtaining the desired ground truth magnitudes, but would still capture the overall spatial trends in vertical winds.     

\begin{figure}[!t]
\vspace{-0.cm}\begin{center}
\begin{tabular}{cc||cc}
\multicolumn{2}{c}{\footnotesize  {layer $k$ at 900-850 hPa  }}&\multicolumn{2}{c}{\footnotesize  {layer $k$ at 700-600 hPa}}\\
\hline \\
\hspace{-0.8cm}\rotatebox{90}{\parbox{2mm}{{{\footnotesize{truth$\,$(ECMWF$\,$proxy)}}}}}\hspace{0.25cm}\includegraphics[width=0.23\textwidth]{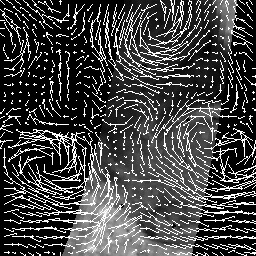}&\hspace{-0.25cm}\includegraphics[width=0.23\textwidth]{./Figures/AMVtrue/W_synth_000006.jpg}&\hspace{-0.cm}\includegraphics[width=0.23\textwidth]{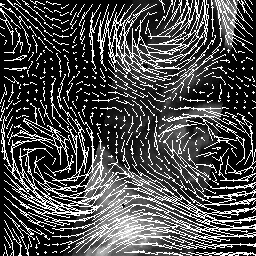}&\hspace{-0.25cm}\includegraphics[width=0.23\textwidth]{./Figures/AMVtrue/W_synth_000002.jpg}\\
\hspace{-0.75cm}\rotatebox{90}{\parbox{2mm}{{{$\quad\quad\quad$\footnotesize{2D}}}}}\hspace{0.25cm}\includegraphics[width=0.23\textwidth]{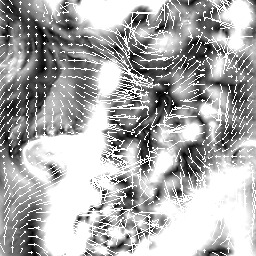}&&\hspace{-0.cm}\includegraphics[width=0.23\textwidth]{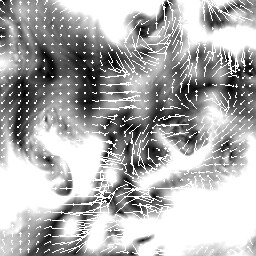}&\hspace{-0.25cm}\\
\hspace{-0.75cm}\rotatebox{90}{\parbox{2mm}{{{$\quad\quad\quad$\footnotesize{3D}}}}}\hspace{0.25cm}\includegraphics[width=0.23\textwidth]{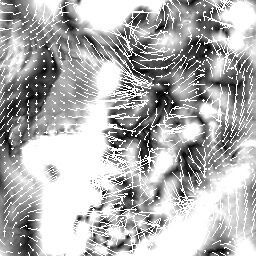}&\hspace{-0.25cm}\includegraphics[width=0.23\textwidth]{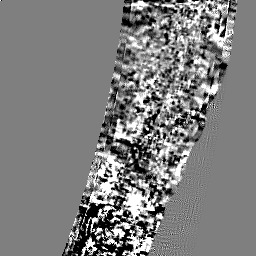}&\hspace{-0.cm}\includegraphics[width=0.23\textwidth]{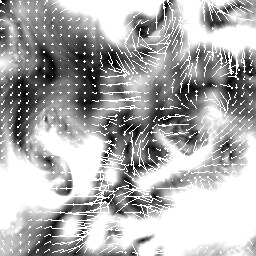}&\hspace{-0.25cm}\includegraphics[width=0.23\textwidth]{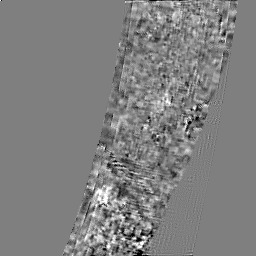}\\
\hspace{-0.75cm}\rotatebox{90}{\parbox{2mm}{{{$\quad\,\,$\footnotesize{3D$\,$hydro.$\,$(hard)}}}}}\hspace{0.25cm}\includegraphics[width=0.23\textwidth]{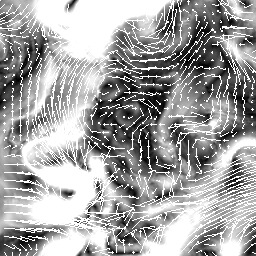}&\hspace{-0.25cm}\includegraphics[width=0.23\textwidth]{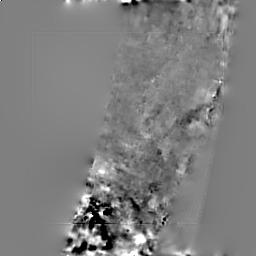}&\hspace{-0.cm}\includegraphics[width=0.23\textwidth]{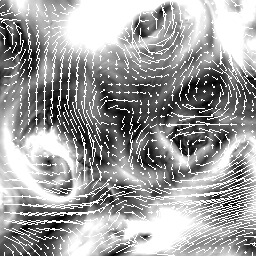}&\hspace{-0.25cm}\includegraphics[width=0.23\textwidth]{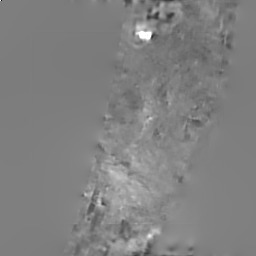}\\
{{\footnotesize{$ {\mathbf{d}^k}^{\star}$ plotted on $\mathbf{y}^{k}_{t_0}$ (above)  }}}&{{\footnotesize{$ {{\boldsymbol{\omega}^k}^\star}$ (above) or $  \boldsymbol{\hat \omega}^{k}$ }}}&{{\footnotesize{$ {\mathbf{d}^k}^{\star}$ (above) or $ {\mathbf{\hat d}^k}$ }}}&{{\footnotesize{$ {{\boldsymbol{\omega}^k}^\star}$ (above) or $  \boldsymbol{\hat \omega}^{k}$ }}}\\
{{\footnotesize{and $ {\mathbf{\hat d}^k}$ plotted on error }}}&&{{\footnotesize{plotted on error norm}}}&
\vspace{-0.cm}
\end{tabular}
	\caption{{\footnotesize  {Horizontal and vertical  components of AMV fields estimated from the IASI data at two different altitudes.  Estimated vector fields  $ {\mathbf{\hat d}^k}$ for the 2D, 3D or 3D hydrostatic models (three last rows) are superimposed on their related map of horizontal EPE whose $j$-th component is $\| \mathbf{d^k}^\star(j)-   \mathbf{\hat d^k}(j) \|_2$.  The ground truth vector fields, the proxy ECMWF horizontal  AMVs fields $ {\mathbf{d}^k}^{\star}$ (row above) , are superimposed on the related humidity field component of noisy and incomplete   observations $\mathbf{y}^{k}_{t_0}$. \label{fig:4}}}}\vspace{-0.cm}
		\end{center}\vspace{-0.75cm}
\end{figure}

We conclude by the quantitative  analysis in Figure~\ref{fig:4} of AMV estimates in the context of the IASI observations. We immediately understand the challenge of extracting 3D AMVs from this noisy and incomplete data set. More precisely, the 2D and 3D models are only partially able to capture the main vortex structures of the horizontal flow located in the vicinity of the observations. The introduction of an hydrostatic balance constraint leads to a dramatic improvement since most of the vortices are accurately estimated in the observation areas but also extrapolated outside these areas.  The vertical component of the AMVs follows the same trends as for the estimates obtained in the previous experiment with ECMWF data, with the difference that the spatial structures are estimated in the case of IASI data only in areas with consecutive observations and that the vertical wind structures are corrupted by a stronger noise.

\section{Conclusion and Perspectives}
Based on atmospheric thermodynamics and hydrostatic equilibrium,  this work proposes a method for estimating vertical profiles of 3D AMV fields from incomplete image observations collected by hyperspectral sounders. The contributions are:  {\it i)}  to devise a geophysically-sound inverse model for optic flow  taking the form of a constrained energy minimization,  {\it ii)} to design an efficient and low complexity algorithm to solve this difficult problem.

The energy to be minimized relies on adiabatic atmospheric thermodynamics. More precisely, the energy measures discrepancies from the layered transport of pressure-averaged temperature, humidity, and ozone concentration fields driven by the action of pressure-averaged horizontal winds, interacting with each other through the vertical winds at the vertical layer boundaries. The AMV fields should {\it in sus} be in hydrostatic balance: the pressure-averaged horizontal divergence of the flow must compensate the difference between vertical winds at the layer lower and upper boundaries. 

AMV estimation from the incomplete observations takes the form of a linear constrained non-convex and high-dimensional energy minimization problem,  where the variables to be jointly optimized are the profile of 3D AMV fields together with the profile of temperature, humidity, and ozone concentration fields. By splitting the global estimation problem into a set of independent optic flow subproblems, and adding a set of matching constraints forcing the split variables to coincide, we take advantage of an ADMM framework to instead solve a set of independent lower-dimensional unconstrained and differentiable minimization problems. The non-convexity of these subproblems is efficiently handled using a multiresolution wavelet expansion of AMVs combined with an L-BFGS optimization scheme.  The resulting global algorithm has advantageous complexity independent of the number of layers and log  linear in the number of image pixels. 

Thanks to the availability of real hyperspectral observations by IASI on Metop satellites and synchronized real NWP provided by ECMWF, we show  in our  numerical evaluation that both, the geophysical modeling and the optimization are quantitatively and qualitatively relevant. In particular, we find a decreases of the estimation error reaching about $35$\%  in comparison to state-of-the-art. 

Although it is beyond the scope of this paper, a comprehensive evaluation of the capabilities of the proposed algorithm, however, remains to be done before operational use.  An important study that remains to be done is the empirical evaluation of the sensitivity of the estimated AMV profile to the profile of the regularization parameters and the number of layers. Such a study is particularly  crucial in the context of IASI observations in order to calibrate some optimal parameters (in the sense that they minimize the estimation error). In the light of the results obtained on the ECMWF dataset, we expect that such a study will allow a more discriminating performance gain between the 2D and 3D model. Of course, future work will concern the validation of the algorithm on longer test periods.

  \appendix

\section{Atmospheric Conservation Laws}\label{sec:appGeophys}
We present in this section bi-dimensional continuous models issued from the vertical integration in the pressure interval $[p^{k},p^{k+1}]$ of atmospheric conservation laws.

\subsection{Constraints Related to the Kinematic Method}
Interesting models for 3D compressible  atmospheric motion adapted to infrared image measurements
may be derived by using an isobaric coordinate
system $(x,y,p)$. In comparison to standard altimetric coordinates $(x,y,z)$, isobaric coordinates are
advantageous: they enable to handle in a simple manner the compressibility of atmospheric
flows. In this coordinate system,  the pressure function $p$ acts as a vertical coordinate. Note that when pressure is used as the vertical coordinate, horizontal partial derivatives must be evaluated holding $p$ constant. 
The vertical wind 
is defined in the isobaric coordinate system by  $\omega=\frac{d p}{dt}.$
Using the
hydrostatic assumption
$
\frac{\partial p}{\partial z}=-\rho g,
$
 relating pressure to  density  $\rho$ and gravity $g$, the expansion of $\omega$ in the $(x,y,z)$ altimetric coordinate system yields 
 \begin{eqnarray}\label{dpdt}
\omega=\frac{\partial p}{\partial t} +  u   \frac{\partial p }{\partial x }+v   \frac{\partial p }{\partial y }  -\textrm{w}\rho g,
\end{eqnarray}
where partial derivative operators  and the vertical velocity $\textrm{w}$ are in $z$ coordinates.

In the light of~\eqref{dpdt}, it is worth noticing that the mass conservation constraint  \eqref{ICE} constitutes  the kernel of the so-called {\it kinematic method},
applied in meteorology for the recovery of vertical motion\footnote{A good approximation for synoptic-scale motions is to let
$
\omega \simeq  -\textrm{w}\rho g.
$
Thus, using this approximation, vertical
motion can be expressed as:
\begin{eqnarray}\label{kinematic}
\textrm{w}(s^{k+1}) = \frac{\rho(s^{k})\textrm{w}(s^{k})}{\rho(s^{k+1})}-\frac{\delta p^k}{g \rho(s^{k+1})}\nabla_s \cdot (\mathbf{v}^{k}),
\end{eqnarray}
where, in relation to pressure levels  $p^{k}$ and $p^{k+1}$, we have define the isobaric surfaces $s^{k}(x,y)$ and $s^{k+1}(x,y)$ in the altimetric coordinate system $(x,y,z)$.
Vertical wind given by \eqref{kinematic} corresponds exactly to the {\it kinematic} estimate, see~\cite{Holton92}.}
\subsection{Constraints Related to the Adiabatic Method}
Assume we  neglect the diabetic heating and omit the modeling of atmosphere chemistry below the stratosphere, as it is commonly done in atmospheric studies \cite{lahoz2010data}.  Then expressing in the isobaric system the first law of thermodynamics and  the dynamics of ozone both yields transport equations of the form of~\cite{Holton92}
\begin{align}\label{eq:thermo}
 \frac{\partial \tvtt }{\partial t}+ u   \frac{\partial \tvtt }{\partial x }+v   \frac{\partial \tvtt }{\partial y } +\gamma \omega=0,
\end{align}
with respectively $\gamma=( \frac{\partial T}{\partial p}-\frac{1}{c_p\rho})$,  $\gamma=\frac{\partial q }{\partial p }$ and  $\gamma=\frac{\partial o}{\partial p }$ for either the temperature  $\tvtt=T$, the humidity  $\tvtt=q$ or the ozone concentration  $\tvtt=o$ variables.  $c_p$ denotes   specific heat of dry air at constant pressure.
Estimation of vertical winds using \eqref{eq:thermo} for  temperature  $\tvtt=T$  is known as the \textit{adiabatic method}.\footnote{ Using natural coordinates, vertical
motion is
$
\textrm{w} =  \frac{1}{(\frac{g}{c_p}+\frac{\partial T}{\partial z})}(\frac{\partial T}{\partial t}+ u   \frac{\partial T }{\partial x }+v   \frac{\partial T }{\partial y }),
$
where dry adiabatic lapse rate  $\frac{\partial T}{\partial z}$ can be approximated by a constant value throughout the lower atmosphere~\cite{Holton92}.}
We are  interested in integrating \eqref{eq:thermo} vertically to relate the evolution of  pressure-averaged temperature fields to  pressure-averaged horizontal displacements. In this perspective, we assume that horizontal displacements  in the pressure  interval $[p^{k},p^{k+1}]$ are  equal to pressure-average horizontal displacements, \textit{i.e.},  $(u,v)\simeq\mathbf{v}^k$ in $[p^{k},p^{k+1}]$. 
Approximating the vertical integration of the last term of \eqref{eq:thermo} using the trapezoidal rule, and inverting the derivatives and integrals (assuming smoothness conditions) in the remaining terms of \eqref{eq:thermo}, we obtain by the vertical integration of~\eqref{eq:thermo} (one of the three components of) the partial differential equation \eqref{eq:OFC}. The  components of  $\gamma^k$ are identified to either  
$ \frac{\partial T}{\partial p}(p^k)-\frac{1}{c_p\rho(p^k)},$  $\frac{\partial q }{\partial p }(p^k)$ or  $\frac{\partial \oo  }{\partial p }(p^k).$
We note that the trapezoidal rule used for vertical integration has been chosen for consistency between the model variables:  the vertical integrated dynamics depend on   vertical winds at the layers interfaces, as for  the mass conservation constraint  \eqref{ICE}.

\section{ ADMM Solver}

\subsection{The ADMM Generic Procedure}\label{app:A}
The alternating direction method of multipliers (ADMM) focusses on the following type of optimization problems:
\begin{align}\label{eq:ADMMstdproblem}
%\begin{array}{l}
\min_{\zz_1\in\Zc_1,\zz_2\in\Zc_1} \mathcal{G}_1(\zz_1)+\mathcal{G}_2(\zz_2),\quad 
\mbox{ s.t.  $\Am\zz_1=-\Bm \zz_2$},
%\end{array}
\end{align}
where 
%%$\x\in\Rb^{aaa}$, $\z\in\Rb^{aaa}$, $\A\in\Rb^{aaa\times bob}$ and 
 $\Am\in\Rb^{r\times n_1}$, $\Bm\in\Rb^{r\times n_2}$,   $\mathcal{G}_1:\Rb^{n_1}\rightarrow\Rb$, $\mathcal{G}_2:\Rb^{n_2}\rightarrow\Rb$ are closed, proper and convex functions, and $\Zc_1$, $\Zc_2$ are nonempty convex sets.
ADMM is an iterative procedure inspired by the well-known method of multipliers \cite{Bertsekas99}. It searches for a minimizer of \eqref{eq:ADMMstdproblem} by sequentially minimizing the corresponding augmented Lagrangian with respect to each primal variables $\zz_1$ and $\zz_2$, before updating a dual variable $\mathbf{u}\in\Rb^r$. Formally, the ADMM recursions take the form:% the following recursion:
\begin{align}
%\begin{aligned}%{l}
\zz_1^{(i+1)}&=\argmin_{\zz_1\in\Zc_1} \mathcal{G}_1(\zz_1)+\frac{\rho}{2} \|\Am\zz_1+\Bm \zz_2^{(i)}+\uv^{(i)}\|^2_2,\label{eq:ADMMstdformA}\\[0.1cm]
\zz_2^{(i+1)}&=\argmin_{\zz_2\in\Zc_2} \mathcal{G}_2(\zz_2)+\frac{\rho}{2} \|\Am\zz_1^{(i+1)}+\Bm \zz_2+\uv^{(i)}\|^2_2,\label{eq:ADMMstdformB}\\[0.1cm]
\uv^{(i+1)}&=\uv^{(k)}+\Am\zz_1^{(i+1)}+\Bm\zz_2^{(i+1)},\label{eq:ADMMstdformC}
%\end{aligned}
\end{align}
for some $\rho>0$. 

%ADMM has recently sparked a surge of interest in the signal-processing community for several reasons. First, the conditions on $\mathcal{G}_1$ and $\mathcal{G}_2$ in \eqref{eq:ADMMstdproblem} (\ie, closed, proper and convex) are mild and \eqref{eq:ADMMstdproblem} therefore encompasses a large number of optimization problems as particular cases. Second, the ADMM recursion \eqref{eq:ADMMstdformA}-\eqref{eq:ADMMstdformC} converges to a solution of \eqref{eq:ADMMstdproblem} under very general conditions, see \cite[section 3.2]{Boyd11}. Third, although ADMM is known to be slow to converge to a solution with high accuracy, it has been shown empirically that ADMM converges to modest accuracy in a few tens of iterations. 
%Finally, the optimization problems involved in the updates of $\x^{(k+1)}$ and $\z^{(k+1)}$ in \eqref{eq:ADMMstdproblem} admit very fast implementation or even closed-form solution in many setups; moreover ADMM is still ensured to converge if inexact minimizations are carried out in the $\x$ and $\z$ updates, see \cite{Eckstein1992DouglasRachford}. These two last features make ADMM very appealing in large-scale problems where modest accuracy is often sufficient but computational load is of utmost importance. 

%\begin{pmatrix} \mathbf{D}&- \mathbf{L}&\mathbf{0}\\  \mathbf{0} & \mathbf{0} &\mathbf{I} \end{pmatrix}\begin{pmatrix}{\mathbf{d}}\\ \boldsymbol{\omega}    \\\boldsymbol{c}\end{pmatrix} +
% \begin{pmatrix}   \mathbf{0}\\  - \mathbf{I}\end{pmatrix} \boldsymbol{\tilde c}  =
% \begin{pmatrix} \mathbf{0}\\ \mathbf{0} \end{pmatrix} .
 
\subsection{Particularization to Problem \eqref{eq:ADMMProb}}\label{app:B}
To solve the non-differentiable non-convex problem~\eqref{eq:ADMMProb}, we use the formalism exposed in  appendix \ref{app:A}  with 
$\zz_1=\begin{pmatrix}{\boldsymbol{d}}\\ \boldsymbol{\omega} \\ \boldsymbol{c}\end{pmatrix}$, $\zz_2= \boldsymbol{\tilde c}$
, $\Zc_1=\Rr^{2Km}\times\Rr^{(K-1)m}\times\Rr^{3Km} $,  $\Zc_2=\Rr^{3Km} $,  $\mathbf{A}=\begin{pmatrix} \mathbf{D}&- \mathbf{L}&\mathbf{0}\\  \mathbf{0} & \mathbf{0} &\mathbf{I} \end{pmatrix} \in \Rr^{  4Km \times (6K -1)m }$ and $ \mathbf{B}= \begin{pmatrix}   \mathbf{0}\\  - \mathbf{I}\end{pmatrix} \in \Rr^{4Km \times 3Km}$. We obtain the  ADMM optimization steps~\eqref{eq:step1_ADMM} and~\eqref{eq:step3_ADMM_prox}, complemented by
$
 \boldsymbol{\tilde c}^{(i+1)} = \argmin_{\boldsymbol{\tilde c}}  \|  \boldsymbol{\tilde c}  \|_1 + \frac{\rho}{2 \boldsymbol{\alpha}_{\xx}} \|\boldsymbol{ c}^{(i+1)} - \boldsymbol{\tilde c}  + \mathbf{u}_{\boldsymbol{c}}^{(i)} \|^2_2.$
The optimization problem specified in the latter has a very simple analytical solution. In fact the right-hand side corresponds to the definition of the proximal operator of the $\ell_1$ norm.  The latter has been extensively studied in the literature  (see \eg, \cite[section 6.5.2]{OPT-003}) and possesses the simple analytical solution~\eqref{eq:step2_ADMM_prox} based on soft-thresholding operators~\eqref{eq:softthresh}.

\subsection{Particularization to Problem~\eqref{eq:ADMMProbSplit}}\label{app:Bbis}
We split the set of optimization variables, except $\boldsymbol{\omega}$,  into the two subsets 
$
\{\mathbf{d},\boldsymbol{c},\boldsymbol{\tilde c}\}=\{\mathbf{d}',\boldsymbol{c}',\boldsymbol{\tilde c'}\}\cup\{\mathbf{d}^{''},\boldsymbol{c^{''}},\boldsymbol{\tilde c^{''}}\},
$
where the subscripts $\cdot^{'}$ and $\cdot^{''}$ denote  the subset of variables indexed respectively by height level number $k$ in $\mathcal{K}'=\{0,2,4,\ldots,K-2\}$ and $\mathcal{K}''=\{1,3,5,\ldots,K-1\}$.
Using this splitting, problem~\eqref{eq:ADMMProbSplit} is  rewritten as
 \begin{equation}
\left\{\begin{aligned}
%&(\mathbf{e}_{\boldsymbol{  d}} , \boldsymbol{\omega},\boldsymbol{\tilde \omega},\boldsymbol{h}, \boldsymbol{\tilde h},\boldsymbol{c},\boldsymbol{\tilde c},\boldsymbol{\epsilon}) 
%=\\
& \argmin_{\mathbf{d},\boldsymbol{\omega},\boldsymbol{\tilde \omega},\boldsymbol{c},\boldsymbol{\tilde c}} \mathcal{J}^{'}(\mathbf{d}^{'}, \boldsymbol{\omega},\boldsymbol{c}^{'})+\mathcal{J}^{''}(\mathbf{e}_\mathbf{d}^{''}, \boldsymbol{\tilde \omega},\boldsymbol{c}^{''})+ \boldsymbol{\alpha_d}\mathcal{R}_d({\boldsymbol{d}} )+ \boldsymbol{\alpha}_{\xx}\mathcal{R}_\xx(\boldsymbol{\tilde c}),\\
&\textrm{s.t.}\quad 
\begin{pmatrix} \mathbf{D}&- \mathbf{L}&\mathbf{0}\\  \mathbf{0} & \mathbf{I} & \mathbf{0} \\  \mathbf{0} & \mathbf{0} &\mathbf{I} \end{pmatrix}\begin{pmatrix}{\boldsymbol{ d}}\\ \boldsymbol{\omega}    \\\boldsymbol{c}\end{pmatrix} 
+ \begin{pmatrix}  \mathbf{0}& \mathbf{0}\\ -\mathbf{I}&  \mathbf{0}\\  \mathbf{0}&  -\mathbf{I}\\ \end{pmatrix}\begin{pmatrix} \boldsymbol{\tilde \omega}    \\ \boldsymbol{\tilde c} \end{pmatrix} 
=\begin{pmatrix} \mathbf{0}\\ \mathbf{0}\\ \mathbf{0}\end{pmatrix} ,
  \end{aligned}\right.
\end{equation}
where $\mathcal{J}=\mathcal{J}^{'}+\mathcal{J}^{''}$ is a decomposition of the cost $\mathcal{J}$ into  two sub-costs $\mathcal{J}^{'}$ and $\mathcal{J}^{''}$,  defined by  the squared $\ell_2$ norm of the vector  of components $\{{\boldsymbol{\delta}_{t}^{k,\ell,s}}(\boldsymbol{\theta},\yy)\}_{k,\ell,s,t}$ defined in~\eqref{eq:defDeltax} with  index $k$ in respectively $\mathcal{K}'$ and $\mathcal{K}''$ (and the other indices $\ell\in\{1,2,3\}$,$s:\chi(s)\in \Omega_m$ and $t\in \{t_0,t_1\}$).
 We then use the formalism exposed in  appendix \ref{app:A}  with
 
$\zz_1=\begin{pmatrix}{\boldsymbol{d}}'\\ \boldsymbol{\omega} \\ \boldsymbol{c}' \\  \boldsymbol{\tilde c}^{''}\end{pmatrix}$, 
$\zz_2=\begin{pmatrix}{\boldsymbol{d}}^{''}\\ \boldsymbol{\tilde \omega} \\ \boldsymbol{\tilde c}' \\ \boldsymbol{c}^{''}\end{pmatrix}$,
$\mathbf{A}= \begin{pmatrix} \mathbf{D}&- \mathbf{L}&\mathbf{0}&\mathbf{0}\\
  \mathbf{0} &  \mathbf{I} &\mathbf{0} & \mathbf{0}\\
    \mathbf{0} & \mathbf{0} & \mathbf{I} & \mathbf{0} \\
      \mathbf{0} & \mathbf{0} & \mathbf{0} & -\mathbf{I}\end{pmatrix}$, 
$ \mathbf{B}=  \begin{pmatrix} \mathbf{D}&- \mathbf{L}&\mathbf{0}&\mathbf{0}\\
  \mathbf{0} & - \mathbf{I} &\mathbf{0} & \mathbf{0}\\
    \mathbf{0} & \mathbf{0} &- \mathbf{I} & \mathbf{0}\\
      \mathbf{0} & \mathbf{0} & \mathbf{0} & \mathbf{I}\end{pmatrix}$, 
      $\Zc_1=\Rr^{Km}\times\Rr^{(K-1)m}\times\Rr^{\frac{3Km}{2}} \times\Rr^{\frac{3Km}{2}}  $ , $\Zc_2=\Zc_1$, with  $\mathbf{A},\mathbf{B}\in\Rr^{(4K-1)m  \times (5K-1)m }$, and where with an  abuse of notations, we have denote identically operator  $\mathbf{D}$ and  $\mathbf{L}$ when they are defined for the two  subsets of variables. We obtain the following ADMM recursions: 
\begin{align}
 &\left\{
 \begin{array}{ll}
&({\mathbf{d}'}^{(i+1)},\boldsymbol{\omega}^{(i+1)}, \boldsymbol{c'}^{(i+1)} ) \\
&\quad \quad \quad=\underset{{\boldsymbol{d}'},\boldsymbol{\omega},\boldsymbol{c'}}{\argmin}\,  \mathcal{J}'(\mathbf{d}', \boldsymbol{\omega}, \boldsymbol{c'})   + \sum_{k\in \mathcal{K}'}\frac{\alpha_d^k}{2}\|{\boldsymbol{\Delta}_m} \,{{\mathbf{d}'}}^k\|^2_2+ \frac{\rho}{2} \|  \mathbf{c'} -   \mathbf{\tilde c'} \,^{(i)}+  \mathbf{u}_{\mathbf{c'}}\,^{(i)} \|^2_2 \\
&\quad \quad \quad\quad \quad \quad\quad \quad \quad+ \frac{\rho}{2 } \|\boldsymbol{ \omega} - \boldsymbol{\tilde \omega}^{(i)}  + \mathbf{u}_{\boldsymbol{\omega}}^{(i)} \|^2_2 
%& \hspace{3cm} +\gamma\, \mathcal{C}(\boldsymbol{\omega}-  \boldsymbol{\omega} ^{(j)},\boldsymbol{h}-  \boldsymbol{h} ^{(j)},\boldsymbol{q}-  \boldsymbol{q} ^{(j)},\boldsymbol{T}-  \boldsymbol{T} ^{(j)},\boldsymbol{o}-  \boldsymbol{o} ^{(j)})  \\
+\frac{\rho}{2}\| \mathbf{D}\mathbf{d}'-\mathbf{L}\,\boldsymbol{\omega} + \mathbf{u}_{\mathbf{{\mathbf{d}}^{'}}}^{(i)} \|^2_2  \\
&\boldsymbol{\tilde c^{''}}^{(i+1)} = \underset{\boldsymbol{\tilde c^{''}}}{\argmin}  \|  \boldsymbol{\tilde c^{''}}  \|_1 + \sum_{k\in \mathcal{K}''} \frac{\rho}{2 {\alpha}^k_{\xx}} \|{\boldsymbol{ c^{''}}^k}^{(i+1)} - {\boldsymbol{\tilde c^{''}}}^k  + \mathbf{u}_{{\boldsymbol{c}^{''}}^k}^{(i)} \|^2_2 ,\label{eq:step1_ADMMSplit} \\
 \end{array}
 \right.
 \end{align}
 \begin{align}
 &\left\{
 \begin{array}{ll}
&({\mathbf{d}''}^{(i+1)},\boldsymbol{\tilde \omega}^{(i+1)}, \boldsymbol{c''}^{(i+1)} ) \\
&\quad \quad \quad=\underset{{\boldsymbol{d}''},\boldsymbol{\tilde \omega},\boldsymbol{c''}}{\argmin}\,  \mathcal{J}''(\mathbf{d}'', \boldsymbol{\tilde \omega}, \boldsymbol{c''})   + \sum_{k\in \mathcal{K}''}\frac{\alpha_d^k}{2}\|{\boldsymbol{\Delta}_m} \,{{\mathbf{d}''}}^k\|^2_2+ \frac{\rho}{2} \|  \mathbf{c''} -   \mathbf{\tilde c''} \,^{(i)}+  \mathbf{u}_{\mathbf{c''}}\,^{(i)} \|^2_2 \\
&  \quad \quad \quad\quad \quad \quad\quad \quad \quad+  \frac{\rho}{2 } \|\boldsymbol{ \omega}^{(i+1)} - \boldsymbol{\tilde \omega}  + \mathbf{u}_{\boldsymbol{\omega}}^{(i)} \|^2_2  
%& \hspace{3cm} +\gamma\, \mathcal{C}(\boldsymbol{\omega}-  \boldsymbol{\omega} ^{(j)},\boldsymbol{h}-  \boldsymbol{h} ^{(j)},\boldsymbol{q}-  \boldsymbol{q} ^{(j)},\boldsymbol{T}-  \boldsymbol{T} ^{(j)},\boldsymbol{o}-  \boldsymbol{o} ^{(j)})  \\
 +\frac{\rho}{2}\| \mathbf{D}{\mathbf{d}^{''}}-\mathbf{L}\,\boldsymbol{\tilde \omega} + \mathbf{u}_{\mathbf{{\mathbf{d}}}^{''}}^{(i)} \|^2_2 \\
&\boldsymbol{\tilde c^{'}}^{(i+1)} = \underset{\boldsymbol{\tilde c^{'}}}{\argmin}  \|  \boldsymbol{\tilde c^{'}}  \|_1 + \sum_{k\in \mathcal{K}'} \frac{\rho}{2 {\alpha}^k_{\xx}} \|{\boldsymbol{ c^{'}}^k}^{(i+1)} - {\boldsymbol{\tilde c^{'}}}^k  + \mathbf{u}_{{\boldsymbol{c}^{'}}^k}^{(i)} \|^2_2 ,
%& \hspace{3cm} +\gamma\, \mathcal{C}(\boldsymbol{\omega}-  \boldsymbol{\omega} ^{(j)},\boldsymbol{h}-  \boldsymbol{h} ^{(j)},\boldsymbol{q}-  \boldsymbol{q} ^{(j)},\boldsymbol{T}-  \boldsymbol{T} ^{(j)},\boldsymbol{o}-  \boldsymbol{o} ^{(j)})  \\
\label{eq:step2_ADMMSplit} \\
 \end{array}
 \right.\\
 &\left\{
 \begin{array}{ll}
\mathbf{u}_{\boldsymbol{\omega}}^{(i+1)} =& \mathbf{u}_{\boldsymbol{\omega}}^{(i)} + \boldsymbol{\omega}^{(i+1)} -\boldsymbol{\tilde \omega}^{(i+1)}, \\
\mathbf{u}_{\mathbf{{\mathbf{d}}^{'}}}^{(i+1)}=& \mathbf{u}_{\mathbf{{\mathbf{d}}^{'}}}^{(i)} + \mathbf{D}{\mathbf{d}'}^{(i+1)}-\mathbf{L}\, \boldsymbol{ \omega}^{(i+1)}, \\
\mathbf{u}_{\mathbf{{\mathbf{d}}^{''}}}^{(i+1)}=& \mathbf{u}_{\mathbf{{\mathbf{d}}^{''}}}^{(i)} + \mathbf{D}{\mathbf{d}^{''}}^{(i+1)}-\mathbf{L}\, \boldsymbol{\tilde \omega}^{(i+1)}, \label{eq:step3_ADMMSplit}\\
\mathbf{u}_{\boldsymbol{c'}}^{(i+1)} =& \mathbf{u}_{\boldsymbol{c'}}^{(i)} + \boldsymbol{c'}^{(i+1)} -\boldsymbol{\tilde c'}^{(i+1)}, \\
\mathbf{u}_{\boldsymbol{c^{''}}}^{(i+1)} =& \mathbf{u}_{\boldsymbol{c^{''}}}^{(i)} + \boldsymbol{c^{''}}^{(i+1)} -\boldsymbol{\tilde c^{''}}^{(i+1)}.
\end{array}
 \right.
\end{align}

 The first minimization appearing in systems  \eqref{eq:step1_ADMMSplit} are  \eqref{eq:step2_ADMMSplit} involve  $K/2$ independent  optimization procedures detailed in~\eqref{eq:step1_ADMMSplitMore}, which can be  naturally  parallelized. The addition of vectors in the ADMM step  \eqref{eq:step3_ADMMSplit} can also be decomposed as exposed in~\eqref{eq:step3_ADMMSplitMore}. Based on soft-thresholding operators~\eqref{eq:softthresh},  the proximal operators of the $\ell_1$ norm specified in \eqref{eq:step1_ADMMSplit} and  \eqref{eq:step2_ADMMSplit} have  simple analytical forms given in~\eqref{eq:step2_ADMMSplitMore}.

\bibliographystyle{spmpsci}
{
\bibliography{ref,group-15302,cherzet,old}
}

\end{document}